\documentclass[]{jair}

\setcopyright{cc}
\copyrightyear{2026}
\acmDOI{10.1613/jair.1.18587}

\JAIRAE{Erez Karpas}
\JAIRTrack{} 
\acmVolume{85}
\acmArticle{35}
\acmMonth{03}
\acmYear{2026}

\RequirePackage[
  datamodel=acmdatamodel,
  style=acmauthoryear,
  backend=biber,
  giveninits=true,
  uniquename=init
  ]{biblatex}

\usepackage{latexsym}
\usepackage{caption}
\usepackage{graphicx}
\usepackage{amsmath}
\usepackage{amsthm}
\usepackage{booktabs}
\usepackage{enumitem}

\usepackage{algorithm}
\usepackage{algpseudocode}
\usepackage[switch]{lineno}

\usepackage{listings}
\usepackage{longtable}
\usepackage{amsfonts}
\usepackage[T1]{fontenc}
\usepackage{tikz}
\usetikzlibrary{arrows,decorations.pathmorphing,backgrounds,positioning,fit,petri}
\usepackage{subcaption}
\usetikzlibrary{fit}
\usetikzlibrary{tikzmark}
\usepackage{lscape}

\usepackage{pgfplotstable, booktabs}
\usepackage{array, multirow}
\usepackage{xcolor,colortbl}
\usepackage{rotating}
\usepackage{placeins}
\usepackage{float}
\usepackage{url}
\usepackage{nicefrac, xfrac}

\newtheorem{theorem}{Theorem}
\newtheorem{example}{Example}
\newtheorem{definition}{Definition}
\newtheorem{rle}{Rule}

\begin{document}

\title{Improving Plan Execution Flexibility using Block-Substitution}

\author{Sabah Binte Noor}
\authornote{Corresponding Author.}
\orcid{0000-0002-8519-9885}
\email{sabah@duet.ac.bd}
\affiliation{%
  \institution{Dhaka University of Engineering \& Technology}
  \city{Gazipur}
  \country{Bangladesh}
}

\author{Fazlul Hasan Siddiqui}
\orcid{0000-0003-4825-6315}
\email{siddiqui@duet.ac.bd}
\affiliation{%
  \institution{Dhaka University of Engineering \& Technology}
  \city{Gazipur}
  \country{Bangladesh}
  }

\renewcommand{\shortauthors}{Noor \& Siddiqui}

\begin{abstract}
Partial-order plans in AI planning facilitate execution flexibility due to their less-constrained nature. Maximizing plan flexibility has been studied through the notions of plan deordering, and plan reordering. Plan deordering removes unnecessary action orderings within a plan, while plan reordering modifies them arbitrarily to minimize action orderings. This study, in contrast with traditional plan deordering and reordering strategies, improves a plan's flexibility by substituting its subplans with actions outside the plan for a planning problem. Our methodology builds on block deordering, which eliminates orderings in a POP by encapsulating coherent actions in blocks, yielding a hierarchically structured plan termed a Block Decomposed Partial-Order (BDPO) plan. We consider the action blocks in a BDPO plan as candidate subplans for substitutions, and ensure that each successful substitution produces a plan with strictly greater flexibility. In addition, this paper employs plan reduction strategies to eliminate redundant actions within a BDPO plan. We also evaluate our approach when combined with MaxSAT-based reorderings. Our experimental result demonstrates a significant improvement in plan execution flexibility on the benchmark problems from International Planning Competitions (IPC), maintaining good coverage and execution time.
\end{abstract}



\received{23 October 2025}
\received[accepted]{27 February 2026}

\maketitle

\tikzstyle{arrow} = [thick,->,>=stealth]
\tikzstyle{process} = [rectangle,rounded corners, thick, minimum width=.6cm, minimum height=.5cm, text centered, draw=black]

\section{Introduction}
An agent must adapt to unanticipated changes to function effectively in a dynamic world. In AI planning, several approaches exist to enhance an agent's flexibility, including allowing the agent to choose from a set of different plans \cite{1decaf} and extending the applications of existing plans through plan generalization \cite{Anderson1988}. The latter approach generates partial-order plans by delaying commitments to the plan's action orderings until necessary. The least-commitment nature of a partial-order plan (POP) enables the potential for parallel execution of actions, reducing the overall completion time. Beyond parallelism, a POP also provides execution flexibility by allowing multiple valid linearizations for the same planning task. This flexibility enables the agent to adjust the execution sequence of actions when unexpected events arise, resources fluctuate, or action durations vary, thereby allowing POPs to be scheduled for improved efficiency or robustness \cite{policella2004}. To maximize a plan's flexibility, prior research explored the notions of plan reordering \cite{maxsat} and deordering \cite{Kambhampati1,Veloso2002,siddiqui_patrik_2012}. Plan deordering eliminates unnecessary action orderings from a plan, while plan reordering allows arbitrary modifications to action orderings. This study focuses on enhancing a plan's flexibility by optimizing resources of the planning problem through substitutions of its action sets (subplans) on top of plan deordering and reordering. 

Partial-order planning can place two actions in a plan without specifying which comes first, reflecting the least commitment strategy.  A sequential plan, in contrast with POP, specifies a total order over a plan's actions. Current heuristic-based forward-search planners can generate sequential plans very efficiently. We can exploit the efficiency of these planners, and attain the advantages of partial-order plans by transforming the sequential plans into POPs. Explanation-based order generalization (EOG) \cite{kk,Veloso2002} deordering techniques can generate a POP from a sequential plan in polynomial time. However, finding the least constrained POP (with minimum action orderings) is an NP-hard problem \cite{backstrom1998}. Therefore, researchers have investigated various deordering and reordering strategies to optimize action orderings in POPs. \citeauthor{siddiqui_patrik_2012} (\citeyear{siddiqui_patrik_2012}) introduce block deordering that eliminates more orderings from a POP by clustering coherent actions into blocks. Neither EOG nor block deordering modifies the actions of the given sequential plan. \citeauthor{maxsat} (\citeyear{maxsat}) employ a partial weighted MaxSAT encoding to optimize plan reordering. Their approach removes redundant actions from the plan but doesn't allow introducing new actions. This encoding has been further extended via action \cite{maxsat_reinst} and variable debinding \cite{Waters_Nebel_Padgham_Sardina_2018} to maximize the flexibility of a partial-order plan. These latter works modify actions by reinstantiating their parameters to optimize resources. However, their encoding does not allow replacing an action with another with a different name, or elimination of redundant actions. 

This paper introduces a novel algorithm that substitutes subplans within a POP to improve plan flexibility. This strategy leverages block deordering in two folds. First, we employ block deordering to remove orderings from a POP by grouping actions into blocks. A POP that incorporates blocks is called a block decomposed partial-order (BDPO) plan. Subsequently, we use these blocks within the BDPO plan as potential subplans for substitutions to improve the plan's flexibility.  We estimate the flexibility of a plan, referred to as \emph{flex}, by the ratio of unordered action pairs to the total action pairs.

\begin{example}
Let us consider \emph{elevator} domain where tasks are to transfer passengers from one floor to another using lifts. The following plan from this domain moves passenger $p1$  from floor $n2$ to $n3$, and passenger $p2$ from floor $n1$ to $n2$ using lift $e1$. \verb|move_up| and  \verb|move_down| actions move a lift up and down by one floor, respectively. On the other hand, actions \verb|board| and \verb|leave| take a passenger in and out of a lift on a specific floor, respectively.

\begin{verbatim}
1 (move_down e1 n3 n2) 
2 (board p1 n2 e1) 
3 (move_up e1 n2 n3)  
4 (leave p1 n3 e1)
5 (move_down e1 n3 n2)
6 (move_down e1 n2 n1)
7 (board p2 n1 e1)
8 (move_up e1 n1 n2)
9 (leave p2 n2 e1)
\end{verbatim}
\begin{figure}[!tb]

    \centering
    \small{

    \scalebox{1}{
    \begin{tikzpicture}[node distance = .75cm]
                \node (init)[process] {Start};
                \node (op11)[process, below of = init]{move\_down e1 n3 n2};
                \node (op1)[process, below of = op11, xshift = -2cm, yshift=-.2cm]{board p1 n2 e1};
                \node (op2)[process, below of = op1]{move\_up e1 n2 n3};
                \node (op3)[process, below of = op2]{leave p1 n3 e1};
                \node (op4)[process, below of = op3]{move\_down e1 n3 n2};
                \node (op5)[process, below of = op11, xshift =2cm,  yshift=-.2cm]{move\_down e1 n2 n1};
                \node (op6)[process, below of = op5]{board p2 n1 e1};
                \node (op7)[process, below of = op6]{move\_up e1 n1 n2};
                \node (op8)[process, below of = op7]{leave p2 n2 e1};
                \node (goal)[process, below of= op8,  xshift =-2cm, yshift=-.1cm] {Finish};
                \draw [arrow] (init) ->(op11);
                 
                \draw [arrow] (op1) ->(op2); 
                \draw [arrow] (op2) ->(op3); 
                \draw [arrow] (op3) ->(op4); 

                \draw [arrow] (op5) ->(op6);
                \draw [arrow] (op6) ->(op7);  
                \draw [arrow] (op7) ->(op8); 
            
                \tikzset{every node/.style={draw,dashed, color=black!80, inner sep=+5pt}}
                 
                \begin{pgfonlayer}{background}
                     \node (fit1) [fit=(op1) (op2) (op3) (op4)] [inner sep= 4pt, rounded corners, fill=black!7, label={[xshift=-12mm,yshift = 9mm]center:{$\mathbf{b_1}$}}] {};
                     \node (fit2) [fit=(op5) (op6) (op7) (op8)] [inner sep=4pt,rounded corners, fill=black!7, label={[xshift=12mm,yshift = 7mm]center:{$\mathbf{b_2}$}}]{};
                \end{pgfonlayer}
                    \draw [arrow] (fit2.south) ->(goal); 
                    \draw [arrow] (fit1.south) ->(goal); 
                    \draw [arrow] (op11) ->(fit1.north);
                    \draw [arrow] (op11) ->(fit2.north);
            \end{tikzpicture}
            }
            }
    \caption{ A block-decomposed partial-order (BDPO) plan where the unordered blocks, $b_1$ and $b_2$, can be executed in any order.}
    \label{fig:dpop}
\end{figure}

Conventional deordering strategies \cite{kk,Veloso2002} can not remove any ordering from this plan. However, block deordering can construct two blocks $b_1$ and $b_2$ over actions 2 to 5 and 6 to 9, respectively. Blocks $b_1$ and $b_2$ have no ordering in the resultant BDPO plan (presented in Figure \ref{fig:dpop}). Therefore, block deordering increases the plan flex from 0 to 0.44 (16 unordered pairs out of total 36 action pairs).
If there exists a second lift $e2$, we can further improve the flexibility of this BDPO plan by substituting blocks. Considering that the lift $e2$ is initially  on floor $n1$, we can replace the block $b_2$ with the following subplan by encapsulating it in another block $b_3$.

\begin{verbatim}
1 (board p2 n1 e2)
2 (move_up e2 n1 n2)
3 (leave p2 n2 e2)    
\end{verbatim}

 \begin{figure}[!bp]
    \centering
    \scalebox{1}{
    \small{
    \begin{tikzpicture}[node distance = .75cm]
                \node (init)[process] {Start};
                \node (op11)[process, below of = init,  xshift = -2cm]{move\_down e1 n3 n2};
                \node (op1)[process, below of = op11, yshift=-.2cm]{board p1 n2 e1};
                \node (op2)[process, below of = op1]{move\_up e1 n2 n3};
                \node (op3)[process, below of = op2]{leave p1 n3 e1};
                \node (op4)[process, dashed, below of = op3]{move\_down e1 n3 n2};
                \node (op5)[process, below of = init, xshift =2cm,  yshift=-1.3cm]{board p2 n1 e2};
                \node (op7)[process, below of = op5]{move\_up e2 n1 n2};
                \node (op8)[process, below of = op7]{leave p2 n2 e2};
                \node (goal)[process, below of= op4,  xshift =2cm] {Finish};

                \draw [arrow] (op1) ->(op2); 
                \draw [arrow] (op2) ->(op3); 
                \draw [arrow] (op3) ->(op4); 
                \draw [arrow] (op5) ->(op7); 
                \draw [arrow] (op7) ->(op8); 
      
                \tikzset{every node/.style={draw,dashed, color=black!80, inner sep=+5pt}}
                 
                \begin{pgfonlayer}{background}
                     \node (fit1) [fit=(op1) (op2) (op3) (op4)] [inner sep= 4pt, rounded corners, fill=black!7, label={[xshift=-12mm,yshift = 8.5mm]center:{$\mathbf{b_1}$}}] {};
                     \node (fit2) [fit=(op5) (op6) (op7) (op8)] [inner sep=4pt,rounded corners, fill=black!7, label={[xshift=12mm,yshift = 4mm]center:{$\mathbf{b_3}$}}]{};
                \end{pgfonlayer}
                    \draw [arrow] (init) ->(op11.north);
                    \draw [arrow] (init) ->(fit2.north);
                    \draw [arrow] (op11) ->(fit1);
                    \draw [arrow] (fit2.south) ->(goal.east); 
                    \draw [arrow] (fit1.south) ->(goal.west); 
            \end{tikzpicture}
            }
            }
    \caption{Substituting block $b_2$ in the BDPO plan, presented in Figure \ref{fig:dpop}, with the block $b_3$ deorders action \texttt{move\_down e1 n3 n2}  and block $b_3$, increasing the plan flex from 0.44 to 0.54.}
    \label{fig:dpop2}
\end{figure}

This substitution allows action \verb|move_down e1 n3 n2| and block $b_3$ to be unordered in the BDPO plan (shown in Figure \ref{fig:dpop2}), improving the flex to 0.54.  In addition, the action \verb|move_down e1 n3 n2| in block $b_1$ now becomes redundant. Removing this action from the BDPO plan further increases the flex to 0.75, and decreases the plan cost to seven (assuming unit cost actions).  
\end{example}

This simple example illustrates how block substitution and deordering can increase plan flexibility and reduce costs by minimizing redundant actions. Removing redundant actions from a plan may result in more ordering constraints due to the compactness of the actions.  Despite the impact on plan flexibility, removing unnecessary actions from a plan can still be beneficial. In this work,  we therefore also eliminate redundant actions from BDPO plans.  

We have conducted a comparative study between the proposed method for improving plan flexibility and MaxSAT reorderings. We also evaluate the effectiveness of our method when combined with MaxSAT reorderings.
Our algorithms are implemented on the code base of the Fast Downward planning system \cite{fd}. We experiment our proposed methodologies with the PDDL benchmark problems from International Planning Competitions (IPC).

\section{Preliminaries}
This section presents the semantics of planning task, partial-order planning, and plan deordering and reordering. We also explain different deordering/reordering strategies related to our work. 

\subsection{Planning Task}
This study considers classical planning in finite domain representation (FDR)  \cite{fd}. FDR describes a planning task using a  set of state variables and their corresponding values. Planning tasks are commonly modeled using planning domain definition language (PDDL) \cite{lipovetzky2019introduction}. PDDL Planning tasks can be automatically converted into FDR \cite{HELMERT2009503}.

\begin{definition}
    \label{def:sas}
    A \textbf{planning task in FDR} is a 4-tuple $\Pi =\langle \mathcal{V}, O, s_0, s_* \rangle$ where:
    \begin{itemize}
        \item $\mathcal{V}$ is a finite set of \textbf{state variables}, each with an associated finite domain $\mathcal{D}_v$. A \textbf{fact} is a pair $\langle v,d\rangle$ with $v\in \mathcal{V}$ and $d\in \mathcal{D}_v$. \\
        A \textbf{state} is a function $s$ defined on $\mathcal{V}$, where for all $v \in \mathcal{V}$, there must be $s(v)\in\mathcal{D}_v$. We often notationally treat a state as a set of facts.  A \textbf{partial state} is essentially the function $s$ but defined only on a subset of $\mathcal{V}$,  denoted as $\mathit{vars}(s)$.
        \item O is a finite set of \textbf{operators}. Each operator $o$ has an associated partial state  $pre(o)$ called its \textbf{precondition}, an associated partial state $\mathit{eff}(o)$ called its \textbf{effect}, and an associated nonnegative number   $cost(o) \in \mathbb{R}^+_0$ called its \textbf{cost}.
        \item $s_0$ is the \textbf{initial state}, 
        \item $s_*$ is a partial state representing \textbf{goal} conditions.
    \end{itemize}
\end{definition}

An operator $o$ is applicable in a state $s$ iff $\mathit{pre}(o) \subseteq s$, and applying $o$ in $s$ yields another state $\hat{s} = apply(o,s)$ in which the value of $v$ becomes $d$ for each $\langle v, d \rangle \in \mathit{eff}(o)$.

A plan $\pi$ is a sequence of operators $\langle o_1, o_2, \dots, o_i ,\dots, o_n \rangle$ and is valid for a planning task $\Pi$ iff,
\begin{enumerate}
    \item $\mathit{pre}(o_1)\subseteq s_0$,
    \item  $\forall i \in \{1,2,\dots,n-1\}$ $\mathit{pre}(o_{i+1}) \subseteq s_i$, where $s_i= apply(o_i,s_{i-1})$, and
    \item $s_* \subseteq s_n$, where $s_n= apply(o_n,s_{n-1})$
\end{enumerate}

In FDR, tasks are grounded to keep the formalism simple \cite{HELMERT2009503}, and ground actions are referred to as operators (Definition \ref{def:sas}). Hence, we use the term \textit{operator} instead of \textit{action} in the rest of this paper. FDR operators do not explicitly provide add or delete effects. An operator $o$  produces a fact $\langle v, d \rangle$ if $\langle v, d \rangle$ belongs to $o$'s effect. On the other hand, $o$ deletes a fact $\langle v, d \rangle$ if $o$ changes the value of $v$ from $d$ to another value. Definition \ref{def:op} formally defines whether a fact is consumed, produced, or deleted by an operator. Notably, this is a simple case, which will be expanded upon later.

\begin{definition}
\label{def:op}
  The set of facts that are consumed, produced, and deleted by an operator $o$ are denoted as \textbf{cons(o)}, \textbf{prod(o)}, and \textbf{del(o)}, respectively.
    \begin{itemize}
        \item A fact $\langle v, d \rangle\in cons(o)$ iff $\langle v, d \rangle \in \mathit{pre}(o)$.
        \item  A fact $\langle v, d \rangle\in prod(o)$ iff $\langle v, d \rangle \in \mathit{eff}(o)$.
         \item  A fact $\langle v, d \rangle \in del(o)$ iff 
            \begin{enumerate}
                \item  Either  $v \notin vars(\mathit{cons}(o))$ or $\mathit{cons}(o)(v) =  d$,  and 
                \item  $\mathit{eff}(o)(v) =d'$ s.t. $\ d' \in (\mathcal{D}_v\setminus \{d\})$.  
            \end{enumerate}
    \end{itemize}
\end{definition}

When an operator $o$ does not consume a fact with variable $v$ (i.e.,  $v \notin vars(\mathit{cons}(o))$) and sets the variable $v$ to a value other than $d$, the state in which $o$ is applied determines whether $o$ deletes  $\langle v, d \rangle$ or not. Since the current state is unavailable,  Definition \ref{def:op} states that $o$ deletes $\langle v, d \rangle$ in this scenario to prevent the possibility of overlooking a potential deleter.

\subsection{Partial-Order Planning}
A partial-order plan (POP) specifies a partial order over plan operators, and allows unordered operators to be executed in any sequence.   Definition \ref{def:pop} defines a POP with respect to a planning task $\Pi=\langle \mathcal{V}, O, s_0, s^*\rangle$.  Though an operator can appear more than once in a POP, the definition assumes that every operator is uniquely identifiable.

\begin{definition}\label{def:pop}
A \textbf{partial-order plan} is a 2-tuple $\pi_{pop} =\langle \mathcal{O}, \prec \rangle$ where:
\begin{itemize}
    \item $\mathcal{O}$ is a set of operators.
    \item $\prec$ is a set of ordering constraints over $\mathcal{O}$. An \textbf{ordering constraint} between a pair of operators, $o_i$ and $o_j$  s.t. $o_i, o_j\in \mathcal{O}$, written as $o_i \prec o_j$, states that the operator $o_i$  must be executed anytime before the operator $o_j$. 
\end{itemize}
\end{definition}
 
The ordering constraint $\prec$ has the transitivity property, meaning if $o_i \prec o_j$ and $o_j \prec o_k$, then $o_i \prec o_k$. An ordering $o_i \prec o_j$ is called a basic ordering if it is not transitively implied by other orderings in $\prec$. A linearization of $\pi_{pop}$ is a total ordering of the operators in $\mathcal{O}$. 

The \emph{producer-consumer-threat} (PCT) formalism \cite{backstrom1998} establishes the ordering structure of a partial-order plan by first identifying which operators produce, consume, or delete which facts, and then specifying operator orderings, called causal links, to map each precondition of an operator to another operator's effect. A threat refers to an operator $o_t$ that deletes a fact, and can be executed between two operators $o_i$ and $o_j$, where there exists a causal link from  $o_i$  to $o_j$ for providing the fact. 

\begin{definition}
A \textbf{causal link} between $o_i$ and $o_j$, written as $o_i \xrightarrow{\langle v, d \rangle} o_j$, specifies that  $o_i \prec o_j$ and the operator $o_i$  provides a fact  $\langle v, d \rangle$ to the operator $o_j$ where $\langle v, d \rangle\in (prod(o_i) \cap cons(o_j))$. 
\end{definition}

\begin{definition}
    A \textbf{threat} represents a conflict between an effect of an operator $o_t$ and a causal link $o_i \xrightarrow{\langle v, d \rangle} o_j$. Operator $o_t$ \textbf{threatens}  $o_i \xrightarrow{\langle v, d \rangle} o_j$ if $o_t$ deletes $\langle v, d \rangle$ and can be ordered between $o_i$ and $o_j$.
\end{definition}

 A threat between an operator $o_t$ and a causal link $o_i \xrightarrow{\langle v, d \rangle} o_j$ can be resolved either by a \textbf{promotion}, adding an ordering constraint $o_t \prec o_i$, or by a \textbf{demotion}, adding an ordering constraint $o_j \prec o_t$. A POP $\pi$ is valid iff every operator precondition is supported by a causal link with no threat \cite{weld_1994}. 

  \citeauthor{siddiqui_patrik_2012} (\citeyear{siddiqui_patrik_2012}) introduce three labels, namely  $PC, CD$, and $DP$, to annotate the ordering constraints in a partial-order plan, defined as follows. 

 \begin{definition} Let $\pi_{pop} =\langle \mathcal{O}, \prec \rangle$ be a partial-order plan, and $Re(o_i \prec o_j)$ be the set of ordering constraints between two operators $o_i$ and $o_j$. $Re(o_i \prec o_j)$ can be formed due to three types of reasons, labeled as \textbf{PC}, \textbf{CD}, and \textbf{DP}.
     \begin{itemize}
    \item PC($\langle v, d \rangle$): Producer-consumer of a fact $\langle v, d \rangle$, $PC(\langle v, d \rangle) \in Re(o_i \prec o_j)$, occurs when  $o_i$ produces  $\langle v, d \rangle$ and $o_j$ consumes $\langle v, d \rangle$. Multiple operators may produce $\langle v, d \rangle$ and similarly, multiple operators can also consume $\langle v, d \rangle$. A causal link assigns one operator $o_i$ to achieve $\langle v, d \rangle$ for the operator $o_j$.
    
    \item CD($\langle v, d \rangle$): Consumer-deleter of a fact $\langle v, d \rangle$, $CD(\langle v, d \rangle) \in Re(o_i \prec o_j)$, occurs when operator $o_i$ consumes the fact   $\langle v, d \rangle$ and $o_j$ also  deletes $\langle v, d \rangle$.
    
    \item DP($\langle v, d \rangle$):  Deleter-producer of a fact  $\langle v, d \rangle$, $DP(\langle v, d \rangle) \in Re(o_i \prec o_j)$,  occurs when an operator $o_i$
    deletes $\langle v, d \rangle$ and there is at least one causal link $o_j \xrightarrow{\langle v, d \rangle} o_k$ for some operator $o_k\in \mathcal{O}$. 
\end{itemize}
\label{def:pc-cd-dp}
 \end{definition}

Here, the label $PC$ signifies the causal links in a POP, while $CD$ and $DP$ represent the plan's demotion and promotion ordering constraints, respectively. These labels help to identify and trace the reasoning behind orderings within a POP.

\subsection{Plan Reordering and Deordering}
Partial-order planning embodies the least commitment strategy, which aims to find flexible plans that allow delaying decisions during plan execution \cite{weld_1994}. Two essential concepts for achieving this flexibility are \emph{plan deordering} and \emph{reordering}. Following \citeauthor{backstrom1998}  (\citeyear{backstrom1998}), we provide the formal definitions of these concepts. It is important to note that the following definitions assume that a POP is transitively closed.
 \begin{definition}
     Let $P=\langle \mathcal{O}, \prec \rangle$ and $Q=\langle \mathcal{O}, \prec' \rangle$ be two partial-order plans for a planning task $\Pi$, then:
     \begin{itemize}
         \item $Q$ is a \textbf{deordering} of $P$  w.r.t. $\Pi$ iff $P$ and $Q$ are both valid POPs and $\prec' \subseteq \prec$.
         \item $Q$ is a \textbf{proper deordering} of $P$  w.r.t. $\Pi$ iff $Q$ is a deordering of $P$ and $\prec' \subset \prec$.
          \item $Q$ is a \textbf{reordering} of $P$  w.r.t. $\Pi$ iff $P$ and $Q$ are both valid POPs.
          \item $Q$ is a \textbf{proper reordering} of $P$  w.r.t. $\Pi$ iff $Q$ is a reordering of $P$ and $\prec' \ne \prec$.
         \item $Q$ is a \textbf{minimum deordering} of $P$  w.r.t. $\Pi$ iff 
         \begin{enumerate}
             \item $Q$ is a deordering of $P$, and
             \item there exists no POP $R=\langle \mathcal{O}, \prec ''\rangle$ s.t. $R$ is a deordering of $P$ and $|\prec''| < |\prec'|$.
         \end{enumerate}
     \end{itemize}
     \begin{itemize}
         \item $Q$ is a \textbf{minimum reordering} of $P$  w.r.t. $\Pi$ iff 
         \begin{enumerate}
             \item $Q$ is a reordering of $P$, and
             \item there exists no POP $R=\langle \mathcal{O}, \prec ''\rangle$ s.t. $R$ is a reordering of $P$ and $|\prec''| < |\prec'|$.
         \end{enumerate}
     \end{itemize}
 \end{definition}
These definitions establish a notion of relative optimality between two POPs based on their orderings. Finding minimum deordering or reordering of a POP is NP-hard and cannot be approximated within a constant factor unless $NP\in DTIME(n^{poly \log n})$ \cite{backstrom1998}.

\subsection{Previous Approaches}

The traditional method for generating a POP involves using partial-order causal link (POCL) planning strategy \cite{weld_1994}. POCL starts with a preliminary POP consisting of an initial and a goal operator. The initial operator has no precondition but has the facts in the initial state as effects, whereas the goal operator has goal conditions as its precondition with no effect. This method incrementally refines this preliminary POP by adding a new operator, imposing ordering constraints between two operators, or creating a causal link between two operators. A POP becomes complete when a causal link with no threat supports every operator precondition.  Every node in a POCL planning search space is a partial-order plan.

Two widely used POCL-based partial-order planners are UCPOP \cite{ucpop} and VHPOP \cite{vhpop}. Partial-order planning, specifically with POCL strategy, also offers a more flexible approach to temporal planning, where the exact placement of operators in time can be postponed until constraints emerge. \citeauthor{Coles_Coles_Fox_Long_2021} (\citeyear{Coles_Coles_Fox_Long_2021}) used the POCL strategy to generate temporal plans using a state-based forward search. They combined the delayed-action ordering commitment of partial-order planning with grounded forward search, guided by a temporally informed heuristic. There are also many POCL-based hierarchical planners \cite{bercher_ijcai2017p68,Bercher_2016,bitmonnot2020fape,bitmonnot:hal-01319768}. Besides POCL, there are other approaches, such as \emph{Petri net unfolding} \cite{petri_net} and \emph{Graphplan} \cite{BLUM1997281}, for generating POPs. Petri net unfolding encodes the evaluation of a forward planning system through the repeated unfolding of a specially designed Petri net, a mathematical structure used to model and analyze the dynamics of discrete distributed systems.  Graphplan generates optimal partial-order plans by using a  compact structure, called \emph{planning graph}, to guide its search for a plan. Later several researchers exploit this graphplan as a preprocessor to other search strategies such as Blackbox \cite{blackbox}, IPP \cite{IPP}, and STAN \cite{STAN}.

An alternative approach for generating POP involves converting a sequential plan into a partial-order plan by deordering or reordering. Some earlier plan deordering strategies generalize and store a sequential plan in triangle tables  \cite{FIKES1971189,Regnier91completedetermination} for plan reuse and modification. Later, triangle tables are used as a pre-process for finding partial orderings from a sequential plan with conditional effect \cite{winner2002analyzing}. Recently, \citeauthor{maxsat} (\citeyear{maxsat}) use partial weighted MaxSAT encoding for optimizing plan flexibility by minimizing orderings in a plan. To further enhance plan flexibility,   \emph{action reinstantiation} \cite{maxsat_reinst,Waters_Nebel_Padgham_Sardina_2018} expanded this MaxSAT encoding by integrating additional formulae, enabling the reassignment of operator parameters. Explanation-based order generalization (EOG), a simple yet powerful strategy for deordering, uses validation structure \cite{KAMBHAMPATI1992193,KAMBHAMPATI1994235,Veloso2002} that acts as proof of the correctness of the plan and adjusts the plan to resolve inconsistencies in that proof. The EOG method has recently been extended to incorporate conditional effects \cite{sabah_siddiqui_2022}. On top of EOG, block deordering \cite{siddiqui_patrik_2012} constructs blocks of coherent operators that allow removing more operator orderings from a partial-order plan. Block deordering is also employed to generate \emph{macro-actions} \cite{fazlul_Chrpa_2015} and optimize plan quality \cite{Siddiqui2015ContinuingPQ}.

The following section delineates reordering and deordering strategies that align with our contributions and experimental studies.

\subsubsection{Partial Weighted MaxSAT-based Reorderings}

The partial weighted maximum satisfiability problem is a variation of the classic SAT problem, distinguishing between \emph{soft} and \emph{hard} clauses. Hard clauses must be satisfied, similar to standard SAT clauses, while soft clauses are optional and carry weights representing their significance. The objective is to find a solution that maximizes the total weight of fulfilled soft clauses while ensuring that all \emph{hard} clauses are satisfied.

\paragraph{MR Encodings}
\citeauthor{maxsat} (\citeyear{maxsat}) encode the problem of finding minimum reordering (MR) of a plan as a partial weighted MaxSAT instance. Given a plan (sequential or partial) $\pi=\langle \mathcal{O}, \prec\rangle$ for a planning task $\Pi$, they encode the problem using three types of propositional variables, and refer to the POP corresponding to the solution as \emph{target} POP.
\begin{itemize}
    \item $x_o$: For every operator $o$ in $\mathcal{O}$, $x_o$  indicates that operator $o$ is in the \emph{target} POP.
    \item $\tau(o_i, o_j)$: For every pair of operators $o_i, o_j$ in $\mathcal{O}$, $\tau(o_i, o_j)$ indicates that operator $o_i$ must precede $o_j$ in the \emph{target} POP.
    \item $\gamma (o_i, \langle v, d \rangle, o_j)$: For every operator $o$ in $\mathcal{O}$, $\langle v, d \rangle\in (cons(o_j) \cap prod(o_i))$, $\gamma (o_i, \langle v, d \rangle, o_j)$ indicates a causal link, $o_i \xrightarrow{\langle v, d \rangle} o_j$ in the \emph{target} POP.
\end{itemize}

In the following formulae, the $\stackrel{k}{(\dots)}$ syntax indicates a soft clause with weight $k$, while no weight marking means a hard clause.
\begin{align}
    & (\neg\tau(o,o))\\
    & \tau(o_i, o_j) \land \tau(o_j, o_k) \xrightarrow{} \tau(o_i, o_k)\\
    & (x_{o_I}) \land (x_{o_G})\\
    & x_{o_i} \xrightarrow{} \tau(o_I, o_i) \land \tau(o_i, o_G)
\end{align}
Formulae 1 and 2 ensure the target POP is acyclic and transitively closed, respectively. Formulae 3 and 4 specify that all operators are ordered after the initial state and before the goal.

\begin{align}
    & x_{o_j} \xrightarrow{} \bigwedge\limits_{\langle v, d \rangle\in cons(o_j)}\ \bigvee\limits_{o_i:\langle v, d \rangle\in prod(o_i)} \tau(o_i, o_j)\land \gamma(o_i, \langle v, d \rangle,o_j)\\
    & \gamma (o_i, \langle v, d \rangle, o_j) \xrightarrow{} \bigwedge\limits_{o_k: \langle v, d \rangle \in del(o_k)} x_{o_k} \xrightarrow{} \tau(o_k, o_i) \lor \tau(o_j, o_k)\\
    &\bigwedge\limits_{\forall o_i, o_j \in \mathcal{O}} \stackrel{1}{\neg\tau(o_i, o_j)}
\end{align}

Formula 5 ensures that a causal link supports each precondition, while Formula 6 ensures these causal links are free from threats, establishing necessary promotion and demotion orderings. Formula 7 introduces a soft clause for negating each ordering. Consequently, a higher weight in an encoding solution leads to fewer orderings within the target POP.

\citeauthor{maxsat} also address the objective of committing as few resources as possible, measured as the sum of operator costs in a plan. To this end, they introduce an extended criterion called the \emph{minimum cost least commitment POP (MCLCP)}.
 \begin{definition}\label{def:mclcp}
     Let $P=\langle \mathcal{O}, \prec \rangle$ and $Q=\langle \mathcal{O}, \prec' \rangle$ be two valid partial-order plans for a planning task $\Pi$, $Q$ is a \textbf{minimum cost least commitment POP (MCLCP)} of $P$  iff  
         \begin{enumerate}
             \item $Q$ is a \textbf{minimum reordering} of $P$  w.r.t. $\pi$ s.t. $\mathcal{O}' \subseteq \mathcal{O}$ and
             \item there exists no valid POP $R=\langle \mathcal{O}, \prec ''\rangle$ for $\Pi$ s.t.  i) $\mathcal{O}'' \subseteq \mathcal{O}$, and ii)  either $cost_{\Pi}(R) < cost_{\Pi}(Q)$, or $cost_{\Pi}(R) = cost_{\Pi}(Q) \land |\prec''| < |\prec'|$
         \end{enumerate}

 \end{definition}
This criterion assumes that every operator in a planning task has a positive cost, and it prioritizes minimizing plan cost over minimizing the number of ordering constraints. To incorporate MCLCP in their encoding, \citeauthor{maxsat} added a soft unit clause, containing the negation of the variable for each operator.

\begin{align}
    & \stackrel{cost_{\Pi}(o)+|\mathcal{O}|^2+1}{(\neg x_{o})}, \forall o \in \mathcal{O}\setminus \{o_I, o_G\}
\end{align}

Here, the weight of any single operator clause exceeds the combined weight of all ordering-constraint clauses, since the total number of ordering constraints can never exceed $|\mathcal{O}|^2$. This weighting ensures that the resulting encodings always yield solutions with minimum plan cost.
\paragraph{MRR Encodings}
\citeauthor{maxsat_reinst} (\citeyear{maxsat_reinst})
extend MR encodings, allowing operators to reinstantiate their parameters for providing additional
flexibility, and refer to their work as minimum reinstantiated reordering (MRR). MRR encodes a partial-order plan as a tuple $P= \langle \mathcal{O}, \theta, \prec\rangle$, where $\mathcal{O}$ is a set of operators, $\theta$ is a ground substitution which is complete with respect to $\mathcal{O}$, and $\prec$ is a strict, transitively closed partial order over $\mathcal{O}$.  They introduce the notion of \emph{reinstantiated reorderings} that allows altering operator parameters along with ordering constraints. Let $P= \langle \mathcal{O}, \theta, \prec\rangle$ and  $Q= \langle \mathcal{O}, \theta', \prec'\rangle$ be two  partial-order plans for a planning task $\Pi$, $Q$ is a \textbf{reinstantiated reordering} of $P$ iff $P$ and $Q$ are both valid. $Q$ is  a \textbf{minimum reinstantiated reordering} of $P$ iff $Q$ is a reinstantiated reordering of $P$ and  there is no POP $R=\langle \mathcal{O}, \theta'', \prec''\rangle$ such that $R$ is a reinstantiated reordering of $P$ and $|\prec''| < |\prec'|$.

MRR formulae use letters such as $x, y$ and $z$ to represent variables, $c$ to represent constants, and $t, u$ and $v$ to denote terms. A term is an ordered list of elements, and $t[i]$ refers to the $i$-th element of the list.  A substitution $\theta$ is a mapping from variables to terms, for example $\theta = \{x_1/t_1, \dots x_n/t_n\}$ maps each variable $x_i$ to $t_i$. In a structure $\eta$, $vars(\eta)$ and $consts(\eta)$ represents the variables and constants in $\eta$.

An operator is represented by a tuple $o=\langle vars(o), pre(o), \mathit{eff}(o)\rangle$, where $vars(o)$ is a list of variables, $pre(o)$ and $\mathit{eff}(o)$ are finite sets of (ground or nonground) facts with variables taken from $vars(o)$. An operator $o$ is ground when $pre(o)$ and $\mathit{eff}(o)$ are sets of ground facts. 

A causal link is written as $\langle o_p, q(\overrightarrow{t}), o_c, q(\overrightarrow{u})\rangle$, where $o_p$ is an operator that produces the fact $q(\overrightarrow{t})$, and $o_c$ is an operator that consumes the same fact  $q(\overrightarrow{u})$(i.e., $q(\overrightarrow{t}) \in prod(o_p)$ and $q(\overrightarrow{u}) \in cons(o_c)$). The notation $q(\overrightarrow{t})$ and $q(\overrightarrow{u})$ suggests that they may involve different terms (variables or constants). A causal link $\langle o_p, q(\overrightarrow{t}), o_c, q(\overrightarrow{u})\rangle$ is valid if (i) $\theta(\overrightarrow{t}) = \theta(\overrightarrow{u})$, and (ii) for all operator $o_t$, $q(\overrightarrow{v})$ s.t. $\neg q(\overrightarrow{t}) \in \mathit{eff}(o_t)$, either $\theta(\overrightarrow{u}) = \theta(\overrightarrow{v})$, $o_t\prec o_p$ or $o_c\prec o_t$.

Along with the propositional variables $x$ and $\tau$ used in MR encodings, MRR introduces two additional types of propositional variables,

 \begin{itemize}
    \item $\kappa(t=u)$: For every pair of  variables/constants $t, u$ in $\theta$, $\kappa(t=u)$ encodes that $\theta(t)= \theta(u)$ in the \emph{target} POP .
    \item  $\rho(o_p, \overrightarrow{t}, o_c, \overrightarrow{u})$: For every operator $o$ in $\mathcal{O}$, $q(\overrightarrow{u}) \in cons(o_c)$ and $q(\overrightarrow{t}) \in prod(o_p)$, $\rho(o_p, \overrightarrow{t}, o_c, \overrightarrow{u})$ indicates a causal link $\langle o_p, q(\overrightarrow{t}), o_c, q(\overrightarrow{u})\rangle$ in the \emph{target} POP.
 \end{itemize}

\citeauthor{maxsat_reinst} provide the following encodings along with formulae 1-4 to find the minimum reinstantiated reordering (MRR) of a POP $P= \langle \mathcal{O}, \theta, \prec\rangle$. 
\begin{align}
   & \kappa(t=u) \leftrightarrow \kappa(u=t) \\
   & \kappa(t=u) \land \kappa(u=v) \rightarrow \kappa(t=v)\\
   & \bigwedge\limits_{x\in vars(\mathcal{O})} (\bigvee\limits_{c\in consts(\mathcal{O})} \kappa(x=c) \land 
   \bigwedge\limits_{\stackrel{c_1,c_2\in consts(\mathcal{O}):} {c_1\ne c_2}} \neg\kappa(x=c_1) \lor \neg\kappa(x=c_2))\\
    & \bigwedge\limits_{q(\overrightarrow{u})\in cons(o_c)} \ \bigvee\limits_{q(\overrightarrow{t})\in prod(o_p)} \rho(o_p, \overrightarrow{t}, o_c, \overrightarrow{u}) \land \tau(o_p,o_c) \\
    & \rho(o_p, \overrightarrow{t}, o_c, \overrightarrow{u}) \rightarrow  \bigwedge\limits_{1\leq i\leq|\overrightarrow{t}|} \kappa(\overrightarrow{t}[i] = \overrightarrow{u}[i]) \land \bigwedge\limits_{\stackrel{q(\overrightarrow{v})\in del(o_t):}{\overrightarrow{t}= \overrightarrow{v}, o_t \ne o_c}} (\tau (o_t, o_p) \lor \tau(o_c, o_t))
\end{align}

Formulae 8 and 9 state that the equality relation is symmetric and transitive for variables and constants, while Formula 10 guarantees that every variable is assigned to only one object. Formulae 11 and 12 encode the validity of the target POP.

Determining whether a POP has a minimum reinstantiated reordering with fewer than $k$ ordering constraints is $NP-$ complete, and finding minimum reinstantiated reordering can not be approximated within a constant factor \cite{maxsat_reinst}.

MRR, similar to our approach, facilitates operator substitutions by rebinding the parameters of operators to improve plan flexibility. One of the limitations of MRR is that it only allows replacement within operators with the same name. In the context of the \emph{Elevator} domain,  MRR can update a \verb|move_up| operator to another \verb|move_up| operator but not to a \verb|move_down| operator. Consequently, MRR cannot replace an operator set with another having different operator names or size.

\subsubsection{Explanation-based Order Generalization}
Explanation-based ordering generalization (EOG) \cite{kk,Veloso2002} is a plan deordering strategy that constructs a validation structure by adding a causal link for every precondition of all operators in a plan and, then resolves threats to the causal links by \textit{promotions} or \textit{demotions}. 

Let $\pi$ be a  sequential plan of a planning task $\Pi= \langle \mathcal{V}, O, s_0, s_*\rangle$.
EOG (Algorithm \ref{alg:kk_algorithm}) employs a common strategy to replicate the initial state and goal conditions of $\Pi$ by adding two extra operators $o_I$ and $o_G$ to  $\pi$,  where $pre(o_I)=\emptyset$, $\mathit{eff}(o_I) = s_0$,  $pre(o_G) = s_*$, $\mathit{eff}(o_G)=\emptyset$,  $o_I \prec o_G$ and for all operators $o \in (\pi\setminus \{o_I, o_G\})$, $o_I \prec o \prec o_G$. Then, it constructs the validation structure in lines 3-8 and resolves threats in lines 9-11 by adding promotion and demotion orderings.  This algorithm binds the earliest producers to causal links for reducing the chance of unnecessary transitive orderings.
 
\begin{algorithm}[!tbp]
    \caption{EOG}
    \label{alg:kk_algorithm}
        \begin{algorithmic}[1]
            \State \textbf{Input:} a valid sequential plan $\pi =\langle o_1, \dots, o_n \rangle$
            \State \textbf{Output:} a valid partial-order plan
            \For {$1 < i \leq n$}   
            \Comment{Constructing validation structure}
                \For{$\langle v, d \rangle \in cons(o_i)$}
                    \State find min $k < i$ s.t.,
                    
                        \quad\quad1. $\langle v, d \rangle\in prod(o_k)$
                        
                        \quad\quad2. there is no $j$ s.t. $k <j <i$ and $\langle v, d \rangle\in  del(o_j)$.
             
                    \State add $o_k \xrightarrow{\langle v, d \rangle} o_i$ to $\prec$
                \EndFor 
            \EndFor
            \ForAll{$o_i,o_j \in \pi$ s.t. $ i < j $ } 
            \Comment{Resolving threats}
                    \State add $\langle o_i \prec o_j\rangle$ to $\prec$ if there exists an operator $o_k$, for which 
                    \Statex \quad\quad one of the following conditions is true,
                 
                    \quad$1.$ $o_k \xrightarrow{\langle v, d \rangle} o_i$ to $\prec$ and $o_j$ deletes the fact $\langle v, d \rangle$
                    
                    \quad$2.$ $o_j \xrightarrow{\langle v, d \rangle} o_k$ to $\prec$ and $o_i$ deletes the fact $\langle v, d \rangle$
            \EndFor
    \end{algorithmic}
\end{algorithm}

\subsubsection{Block Deordering}
Block deordering eliminates ordering constraints in a partial-order plan by clustering coherent operators into blocks, and transforms the POP into a block decomposed partial-order (BDPO) plan \cite{Siddiqui2015ContinuingPQ}. 

A block encapsulates a set of partially ordered operators in a plan, and operators in two disjoint blocks cannot interleave with each other, enabling the unordered blocks to be executed in any order. Blocks can also be nested, i.e., a block can contain one or more blocks, but they are not allowed to overlap. A partial-order plan incorporating blocks is called a block decomposed partial-order (BDPO) plan. Consistent with the definition of a partial-order plan (Definition \ref{def:pop}),  the following definition also assumes that every operator in a BDPO plan is uniquely identifiable.

\begin{definition}
A \textbf{block decomposed partial-order plan} is a 3-tuple  $\pi_{bdp} =\langle \mathcal{O}, \mathcal{B} \prec\rangle$ where $\mathcal{O}$ is a set of operators, $\mathcal{B}$ is a set of blocks,  and $\prec$ is a set of ordering constraints over $\mathcal{O}$. Let $b \in \mathcal{B} $ be a block comprising a set of operators such that for any two operators $o, o' \in b$, where $o \prec o'$, there exists no other operator  $o'' \notin b$ with $o \prec o'' \prec o'$.
If $b_i, b_j \in \mathcal{B}$ are two blocks, then only one of these three relations, $b_i \subset b_j$, $b_j \subset b_i$, $b_i \cap b_j = \emptyset$ can be true. 
\label{def: bdpo}
\end{definition}
A block, like an operator, can be expressed by its precondition and its effects.  A fact $\langle v, d \rangle$ belongs to the precondition of a block $b$ if an operator $o$ in $b$ consumes $\langle v, d \rangle$, and no other operator in $b$ provides $\langle v, d \rangle$ to  $o$. On the other hand, a fact $\langle v, d \rangle$ belongs to a block's effect if an operator $o$ produces $\langle v, d \rangle$ and no other operator in $b$ that follows $o$ modifies the value of $v$.
\begin{definition}
    Let $\pi_{bdp} =\langle \mathcal{O}, \mathcal{B} \prec\rangle$ be a BDPO plan, where $b \in \mathcal{B}$ be a block. The \textbf{block semantics} are defined as,
  \begin{itemize}
        \item A fact  $\langle v, d \rangle \in pre(b)$ iff there is an operator $o \in b$ with $\langle v, d \rangle \in pre(o)$, and $b$ has no other operator $o'$ such that there exists a causal link $o'\xrightarrow{\langle v, d \rangle } o$.
        \item A fact $\langle v, d \rangle \in \mathit{eff}(b)$ iff there exists an operator $o \in b$ with $\langle v, d \rangle  \in \mathit{eff}(o)$, and   no operator $o'\in b$ has an effect $\langle v, d' \rangle$ where $o \prec o'$ and $d' \in (\mathcal{D}_v\setminus \{d\})$.
    \end{itemize}
    \label{def: block-semantics}
\end{definition}
In contrast with an operator,  a block may have multiple facts over one variable as its effects. For instance, if a block $b$ contains two unordered operators $o$ and $o'$ with  $ \langle v, d \rangle \in \mathit{eff}(o)$ and $\langle v, d'\rangle \in \mathit{eff}(o')$, respectively, where $d\ne d'$ then $b$ has both $\langle v, d \rangle$ and $\langle v, d'\rangle $ as its effect. This is essential to identify the facts that a block produces or deletes (Definition \ref{def1}).  Defining when a block consumes, produces, and deletes a fact is crucial to support producer-consumer-threat (PCT) formalism.

\begin{definition}
    \label{def1}
    The set of facts that are consumed, produced, and deleted by a block $b$ are denoted as \textbf{cons(b)}, \textbf{prod(b)}, and \textbf{del(b)}, respectively.
        \begin{itemize}
            \item  A fact $\langle v,d\rangle \in cons(b)$, iff $\langle v,d \rangle \in pre(b)$.
            \item  A fact $\langle v,d\rangle \in prod(b)$, iff 
            \begin{enumerate}
                \item $\langle v,d\rangle \notin cons(b)$,
                \item $\langle v,d \rangle \in \mathit{eff}(b)$, and
                \item $\langle v, d' \rangle \notin \mathit{eff}(b)$  where $d' \in (\mathcal{D}_v\setminus \{d\})$.
            \end{enumerate}
            \item A fact $\langle v,d\rangle \in del(b)$, iff
            \begin{enumerate}
                \item  Either $v \notin \mathit{vars}(\mathit{cons}(b))$ or  $cons(b)(v) = d$, and 
                \item $\langle v, d' \rangle \in \mathit{eff}(b)$, where $d' \in (\mathcal{D}_v\setminus \{d\})$. 
            \end{enumerate}
        \end{itemize}
\end{definition}

For any two blocks $b_i$ and $b_j$ in a BDPO plan, the notation $b_i \prec b_j$ signifies that there exists two operators $o$ and $o'$ such that $o\in b_i$, $o'\in b_j$, and $o\prec o'$. The $PC, CD$, and $DP$ labels (Definition \ref{def:pc-cd-dp})  can also be employed to annotate the orderings between blocks. Now, we give definitions of \emph{candidate producer} and \emph{earliest candidate producers} with respect to a precondition of a block. These concepts are related to establishing causal links in our algorithms.

\begin{definition}\label{def:candidate_producer}
   Let $\pi_{bdp} =\langle \mathcal{O}, \mathcal{B} \prec\rangle$ be a BDPO plan, and  $b_i, b_j \in \mathcal{B}$ be two blocks, where $\langle v, d \rangle \in cons(b_j)$,
   \begin{itemize}
       \item Block $b_i$ is a \textbf{candidate producer} of a fact $\langle v, d \rangle$ for $b_j$ if
       \begin{enumerate}
           \item  $\langle v, d \rangle \in \mathit{eff}(b_i)$,
           \item $b_i \prec b_j$, and there exists no block  $b_k \in \mathcal{B}$ with $ \langle v, d \rangle\in del(b_k)$, where  $b_j \nprec b_k \nprec b_i$. 
       \end{enumerate}
           
     \item Block $b_i$ is the \textbf{earliest candidate producer} of a fact $\langle v, d \rangle$ for $b_j$ if there is no candidate producer $b_k$ of $\langle v, d \rangle$ for $b_j$ such that $b_k \prec b_i$.
    \end{itemize}
\end{definition}

Block deordering takes a sequential plan as input, and produces a valid BDPO plan. It first transforms the sequential plan into a POP $\pi=\langle \mathcal{O}, \prec \rangle$ using EOG. Then it builds an initial BDPO plan $\pi_{bdp} = (\mathcal{O}, \mathcal{B}, \prec)$ simply by adding a block $b=\{o\}$ to $\mathcal{B}$ for each operator $o\in\mathcal{O}$. Also, for every $ o_i \prec o_j$ in $\prec$,  it adds an ordering $b_i \prec b_j$ where $o_i\in b_i, o_j\in b_j$ and $b_i, b_j \in \mathcal{B}$.  Then, block deordering employs the following rule \cite{siddiqui_patrik_2012} to remove further orderings in $\pi_{bdp}$. The terms \emph{primitive} and \emph{compound} blocks specify blocks with single and multiple operators, respectively.  For describing rules and algorithms, we use the term \emph{block} generally to refer to both primitive and compound blocks.

\begin{rle}
    \label{rule_1}
    Let  $\pi_{bdp} =\langle \mathcal{O}, B \prec \rangle$ be a valid BDPO plan, and  $b_i \prec b_j$ be a basic ordering,
    \begin{enumerate}[label=\roman*.]
        \item 
        \label{rule_1a} Let $PC(\langle v, d\rangle) \in Re(b_i\prec b_j)$ be a ordering reason, and $b$ be a block, where $b_i \in b, b_j \notin b$ and $\forall b' \in \{b\setminus b_i\}, b_i \nprec b'$. $PC(\langle v, d\rangle)$ can be removed from $Re(b_i\prec b_j)$ if $\langle v, d\rangle \in pre(b)$ and $\exists b_p \notin b$ such that $b_p$ can establish causal links $b_p \xrightarrow{\langle v, d\rangle} b_j $ and $b_p \xrightarrow{\langle v, d\rangle} b$.
        
        \item \label{rule_1b}  Let $CD(\langle v, d\rangle) \in Re(b_i\prec b_j)$ be a ordering reason, and $b$ be a block, where $b_i \in b, b_j \notin b$ and $b \cap b_j = \emptyset$. Then $CD(\langle v, d\rangle)$ can be removed from $Re(b_i\prec b_j)$ if $b$ does not consume $\langle v, d\rangle$.
        
        \item \label{rule_1c} Let $CD(\langle v, d\rangle) \in Re(b_i\prec b_j)$ be a ordering reason, and $b$ be a block, where $b_i \notin b, b_j\in b$ and $b_i \cap b = \emptyset$. The $CD(\langle v, d\rangle)$ can be removed from $Re(b_i\prec b_j)$  if $b$ does not delete $\langle v, d\rangle$.
        \item\label{rule_1d} Let $DP(\langle v, d\rangle) \in Re(b_i\prec b_j)$ be a ordering reason, and $b$ be a block, where, $b_j \in b$, but $b_i \notin b$. Then $DP(\langle v, d\rangle)$ can be removed from $Re(b_i\prec b_j)$ if $b$ includes all blocks $b'$ such that $b_j \xrightarrow{\langle v, d\rangle} b'$. 
    \end{enumerate}
\end{rle}

To remove a $PC(\langle v, d\rangle)$ reason  from $Re(b_i\prec b_j)$, Rule 1(i) searches for a block $b_c$ such that $b_c \prec b_i$ and $\langle v, d \rangle \in cons(b_c)$. If $b_c$ is found, it forms a block $b$ encapsulating $b_i$, $b_c$, and all the blocks ordered between $b_c$ and $b_i$. Since $b_c$ consumes  $\langle v, d \rangle$, there must be a block $b_p$ such that $b_p \xrightarrow{\langle v, d \rangle} b_c$. Therefore, Rule 1(i) establishes $b_p \xrightarrow{\langle v, d \rangle} b$ and $b_p \xrightarrow{\langle v, d \rangle} b_j$, allowing  $PC(\langle v, d\rangle)\in Re(b_i\prec b_j)$ reason to be removed. To eliminate $CD(\langle v, d\rangle) \in Re(b_i\prec b_j)$,  Rule 1(ii-iii) seeks a block $b_p$ with $\langle v, d\rangle \in prod(b_p)$.  If $b_p$ precedes $b_i$ and then Rule 1(ii) creates a new block $b$ with blocks $b_p$, $b_i$ and every block $b'$ such that $b_p \prec b'\prec b_i$. On the other hand, if $b_p$ follows $b_j$, Rule 1(iii) forms the new block $b$ by encompassing $b_j$, $b_p$, and every block $b'$ s.t. $b_j \prec b' \prec b_p$. For removing $DP(\langle v, d\rangle) \in Re(b_i\prec b_j)$,  Rule 1(iv) forms a new block $b$  that includes $b_j$, every block $b'$ such that $b_j \xrightarrow{\langle v,d \rangle} b'$, and each block $b''$ with $b_i\prec b''\prec b'$. The new block $b$ functions as a barrier against the corresponding deleter.

 Block deordering starts by examining every ordering to eliminate from the top of the initial BDPO plan  in a greedy manner. This process removes an ordering  $b_i \prec b_j$ if it can eliminate all of its ordering reasons by applying the Rule 1. However, if some reasons can not be eliminated, then $b_i \prec b_j$ persists. If an attempt is unsuccessful, the algorithm moves on to the next ordering. If the algorithm successfully eliminates an ordering, it returns the newly generated BDPO plan. Then, the algorithm recommences deordering from the top of the latest BDPO plan. This iterative process continues until no further deordering is possible with the most recent BDPO plan.

Previous studies \cite{siddiqui_patrik_2012,Siddiqui2015ContinuingPQ} have utilized BDPO plans to improve overall plan quality by locally optimizing subplans with off-the-shelf planners. Their approach exploited block deordring to identify candidate subplans, referred to as \emph{windows}, to be replaced with lower-cost subplans with respect to the respective planning task. Each window $w$ is a subsequence of a linearization of a BDPO plan, where blocks are partitioned into the part to be replaced ($w$), and those ordered before ($p$) and after ($q$) that part. They formulate a subproblem for each candidate window, where the initial state is obtained by progressing the original initial state of the planning task through the operators $p$, and the goal state is the result of regressing the original goal of the planning task through the operators $q$. Each successful replacement produces a strictly better (i.e. lower cost) sequential plan, which is then again block deordered for further local improvements. This iterative procedure continues until no more local improvements are possible. In this paper, we propose a similar method for improving plan flexibility by replacing blocks, though with key differences.

The central distinction between window replacement and our proposed block-substitution (presented in section \ref{subsection: block-substitution})  lies in plan representation. Window replacement is carried out on a linearization of a BDPO plan and outputs a sequential plan, while block-substition directly replaces a block within a BDPO plan, yielding another BDPO plan. Though we also create a subproblem for each candidate block in our method (presented in Section \ref{subsection:fibs}) for improving plan flexibility using block-substitution, our formulation differs fundamentally from that of \citeauthor{Siddiqui2015ContinuingPQ}. Their objective is to reduce cost by finding lower-cost solutions to replace candidate windows. By contrast, we aim to construct subproblems for candidate blocks such that, after substitution, the resulting BDPO plan achieves strictly greater flexibility while preserving the plan’s cost bound. 

In our proposed method, we also allow removing redundant operators after the deordering process.  The following section provides a brief overview of previous approaches for eliminating redundant operators from a plan.

\subsection{Eliminating Redundant Operators}
Although automated planning is generally $PSPACE$-complete \cite{bylander_1994}, satisfiable planners can efficiently solve large planning problems. The plans generated by these satisfiable planners often have redundant operators. A subsequence of operators (i.e. a subplan) in a plan is redundant if it can be removed without invalidating the plan, and a plan without redundant subsequence is called \emph{perfectly justified} \cite{Fink_yang1997}. Nonetheless, deciding whether a plan is perfectly justified is $\mathit{NP}$-complete \cite{Fink_yang1997,Nakhost_Müller_2021}.  The problem of eliminating redundant action is formally defined through the notion of \emph{plan reduction}.

\begin{definition}\label{def:plan-deduction}
 Let $\pi$ be a valid plan for a planning task $\Pi$. $\rho$ is a \textbf{plan reduction (PR)} of $\pi$ if $\rho$ is a valid plan for $\Pi$, and a subsequence of $\pi$ with $|\rho| < |\pi|$. 
\end{definition}

\begin{definition}
Let $\pi$ be a valid plan for a planning task $\Pi$ and $\rho$ be a plan reduction of $\pi$. $\rho$ is a \textbf{minimal plan reduction (MPR)} of $\pi$ if there exist no $\rho'$ such that $\rho'$ is a reduction of $\pi$ with $cost(\rho') < cost (\rho)$. 
\end{definition}

Several heuristic methods can efficiently identify redundant actions in plans. Some of the early approaches are \emph{Backward Justification} and \emph{Linear Greedy Justification} by \citeauthor{Fink_yang1997} (\citeyear{Fink_yang1997}). Following them, we give definitions of backward justification and greedy justification.

\begin{definition}
    Let $\pi$ be a valid plan for a planning task $\Pi$ that achieves a goal $s_*$. An operator $o$ in $\pi$ is \textbf{backward justified} if it provides a fact $\langle v,d \rangle$ to $s_*$ or to another backward justified operator in $\pi$.
\end{definition}

\begin{definition}
     Let $\pi$ be a valid plan for a planning task $\Pi$. An operator $o$ in $\pi$ is \textbf{greedily justified} if removing $o$ and all its subsequent operators that depend on $o$  violates the correctness of the plan. An operator $o'$ \textbf{depends} on $o$ if there is a causal link $o\xrightarrow{\langle v,d\rangle} o'$ or there exists another operator $o''$ with $o''\xrightarrow{\langle v,d\rangle} o'$   where $o''$ depends on $o$.
\end{definition}

A plan is backward justified if all its operators are backward justified, where an operator is backward justified if it provides some facts necessary for achieving the goal. In contrast, greedy justification checks each operator to determine if it is \emph{greedily justified} by removing it and all subsequent operators that depend on it, and verifying whether the resulting plan remains valid. Greedy justification was later reinvented under the name \emph{Action Elimination} \cite{Nakhost_Müller_2021}. However, this algorithm removes redundant operator sets as soon as they are identified, without considering their cost. \emph{Greedy Action Elimination (GAE)} \cite{baylo_lukas_2014} is an improvement of this strategy, which identifies all redundant operator sets beforehand, and removes the one with the highest cost. 

\citeauthor{Chrpa_Leo_2012} (\citeyear{Chrpa_Leo_2012}) present another approach that identifies and removes redundant \emph{inverse operators}  in a plan. Inverse operators (Definition \ref{def:inverse_operator}) are pairs of operators whose consecutive execution results in the same state as before their application.  They investigated the conditions under which pairs of inverse operators can be removed from plans while maintaining their validity.   \citeauthor{Med_Chrpa_2022} (\citeyear{Med_Chrpa_2022}) generalize the concept of inverse operators to any operator sequence and referred to them as \emph{action cycles}. Their work enhances the performance of the GAE algorithm by identifying redundant operator sequences in the form of action cycles, and extracting operators that are not part of any set of redundant operators, termed \emph{plan action landmarks}, before applying GAE.

\begin{definition}\label{def:inverse_operator}
Operators $o$ and $o'$ are \textbf{inverse operators} iff  $o$ and $o'$ are consecutively applied in any state $s$, where $o$ is applicable, resulting in a state $s'$ such that $s' \subseteq s$.
\end{definition}

\citeauthor{baylo_lukas_2014} (\citeyear{baylo_lukas_2014}) addressed the problem of minimal plan reduction using MaxSAT. In contrast, \citeauthor{Salerno_2023} (\citeyear{Salerno_2023}) encoded the problem of minimal plan reduction as a planning problem, by formulating a modified planning task for the original planning task and a given plan, where operators can be either retained or skipped while preserving their orderings. The optimal solution of the modified task is a \textit{minimal reduction} of the given plan. We refer to this approach as \emph{MPR} (Minimal Plan Reduction). In our work, we apply both Backward Justification and Greedy Justification to the resulting BDPO plans and compare their performance with MPR.

\algnewcommand{\TRUE}{\textbf{true}}
\algnewcommand{\FALSE}{\textbf{false}}
\algnewcommand{\LeftComment}[1]{\Statex \(\triangleright\) #1}
\section{Improving Execution Flexibility using Block-Substitution}
A block decomposed partial-order (BDPO) plan is a hierarchical structure that encloses subplans in blocks within a partial-order plan.
This work introduces a new concept called block-substitution that allows replacing a block (i.e., subplan) in a BDPO plan.  Our proposed algorithm exploits block-substitution to enhance the flexibility of a POP. 

\subsection{Block-Substitution}\label{subsection: block-substitution}
Block-substitution facilitates substituting a block in a valid BDPO plan with another while preserving plan validity.  The term \emph{original block} refers to the block that is being replaced, while  \emph{substituting block} denotes the block taking its place. A block-substitution process requires forming causal links for the substituting block's precondition, and reestablishing all causal links previously supported by the original block. In addition, this process must resolve any potential threat introduced by this substitution to ensure the plan's validity.

\begin{definition}
    Let $\pi_{bdp}=\langle \mathcal{O}, \mathcal{B}, \prec \rangle$ be a valid BDPO plan with respect to a planning task $\Pi =\langle \mathcal{V}, O, s_0, s_* \rangle$, and $b\in \mathcal{B}$ be a block.  Let $\hat{b}=\langle \hat{O},\hat{\prec}\rangle$ be a partial-order subplan w.r.t. the planning task $\Pi$. A \textbf{block-substitution} of $b$ with $\hat{b}$ yields a  BDPO plan $\pi'_{bdp}= \langle \mathcal{O'}, \mathcal{B'}, \prec'\rangle$, where $b\notin \mathcal{B}'$, $\hat{b}\in \mathcal{B}'$. A block-substitution is valid when $\pi'_{bdp}$ is valid.
\end{definition}

\begin{figure}[!bp]

\begin{subfigure}{.48\columnwidth}
     \centering
        \begin{tikzpicture}[node distance= 1.2cm]
            \node (init)[process]{INIT};
            \node (r)[process, below of= init]{$b_r$};
            \node (i)[process, below of= r]{$b_i$};
            \node (s)[process, below of= i,  xshift=1cm, yshift=-.5cm]{$b_s$};
            \node (x)[process, below of= i, xshift=-1cm]{$b_x$};
            \node (t)[process, below of= x]{$b_t$};
            \node (goal)[process, below of= t,xshift=1cm ]{GOAL};
           
            \draw [arrow, dotted](init) -> (r);
            \draw [arrow](r) -> node[right]{$PC(\langle v_1, d_1\rangle)$} (i);
            \draw [arrow](i) ->node[left]{$PC(\langle v_2, d_2\rangle)$}  (x);
            \draw [arrow](x) -> node[left]{$PC(\langle v_3, d_3\rangle)$} (t);
            \draw [ arrow, dotted ](t) -> (goal);
            \draw [ arrow, dotted](s) ->  (goal);
            \draw [arrow](i) -> node[right]{$CD(\langle v_1, d_1\rangle)$} (s);
        \end{tikzpicture}
        \caption{}
        \label{fig:substitution-example-1a}
\end{subfigure}
\begin{subfigure}{.48\columnwidth}
     \centering
        \begin{tikzpicture}[node distance= 1.2cm]
            \node (init)[process]{INIT};
            \node (r)[process, below of= init]{$b_r$};
            \node (i)[process, below of= r]{$b_i$};
            \node (s)[process, below of= i,  xshift=1.2cm, yshift =-.5cm]{$b_s$};
            \node (x)[process, below of= i, xshift=-1.2cm]{$\hat{b}_x$};
            \node (t)[process, below of= x]{$b_t$};
            \node (goal)[process, below of= t,xshift=1.2cm ]{GOAL};
           
            \draw [arrow, dotted](init) -> (r);
            \draw [arrow](r) -> node[right]{$PC(\langle v_1, d_1\rangle)$} (i);
            \draw [arrow](r) -> node[left]{$PC(\langle v_1, d_1\rangle)$}  (x);
            \draw [arrow](x) -> node[left]{$PC(\langle v_3, d_3\rangle)$} (t);
            \draw [ arrow, dotted ](t) -> (goal);
            \draw [ arrow, dotted](s) ->  (goal);
            \draw [arrow](i) -> node[right]{$CD(\langle v_1, d_1\rangle)$} (s);
             \draw [arrow](x) -> node[left, yshift =-.3cm, xshift=1cm]{$(CD\langle v_1, d_1\rangle)$} (s);

        \end{tikzpicture}
        \caption{}
        \label{fig:substitution-example-1b}
\end{subfigure}
\caption{Substituting a block $b_x$ in (a) a valid BDPO plan $\pi_{bdp}=\langle \mathcal{O},\mathcal{B}, \prec\rangle$  with a block $\hat{b}_x \notin \mathcal{B}$, where $\langle v_1, d_1\rangle \in cons(\hat{b}_x)$ and $\langle v_3, d_3\rangle \in prod(\hat{b}_x)$. (b) This substitution adds two causal links $b_r \xrightarrow{\langle v_1, d_1 \rangle} \hat{b}_x$ and $\hat{b}_x \xrightarrow{\langle v_3, d_3 \rangle} b_t$, and an ordering reason $CD(\langle v_1, d_1 \rangle)$ to $Re(\hat{b}_x \prec b_s)$ for resolving threat, producing a valid BDPO plan where blocks $b_i$ and $\hat{b}_x$ are unordered.  The dotted lines represent ordering (basic or transitive) between two blocks. }
\label{fig:substitution-example-1}
\end{figure}

\begin{example}
Let's consider the BDPO plan $\pi_{bdp}=\langle \mathcal{O},\mathcal{B}, \prec\rangle$ in Figure \ref{fig:substitution-example-1a}, where the precondition of block $b_x$ is supported by the causal link $b_i \xrightarrow{\langle v_2, d_2 \rangle} b_x$, and $b_x$ provides $\langle v_3, d_3 \rangle$ to block $b_t$. Block $b_s$ deletes $\langle v_1, d_1 \rangle$ and threatens $b_r \xrightarrow{\langle v_1, d_1\rangle} b_i$. This threat is resolved by adding $CD\langle v_1, d_1 \rangle$ to $Re(b_i \prec b_s)$.  Let $\hat{b}_x\notin \mathcal{B}$ be a block with $\langle v_1, d_1\rangle \in cons(\hat{b}_x)$ and $\langle v_3, d_3 \rangle \in prod(\hat{b}_x)$.  To substitute $b_x$ with $\hat{b}_x$, it is necessary to establish causal links for the precondition of $\hat{b}_x$ and then reestablish the causal link $b_x\xrightarrow{\langle v_3, d_3\rangle} b_t$ as well. That is why, after excluding $b_x$, causal links $b_r \xrightarrow{\langle v_1, d_1\rangle} \hat{b}_x$ and $\hat{b}_x \xrightarrow{\langle v_3, d_3\rangle} b_t$ are added to the resultant BDPO plan (shown in Figure \ref{fig:substitution-example-1b}). Since block $b_s$ deletes $\langle v_1, d_1\rangle$, $b_s$ is a threat to  $b_r \xrightarrow{\langle v_1, d_1\rangle} \hat{b}_x$. To resolve this threat, an ordering reason $CD(\langle v_1, d_1\rangle)$ is added to $Re(\hat{b}_x \prec b_s)$. Since every precondition of each block is now supported by a causal link, and no threat persists, substituting $b_x$ with $\hat{b}_x$ is successful yielding a valid BDPO plan. Notably, the resultant plan has no ordering between the blocks $b_r$ and $\hat{b}_x$.
\end{example}

Block-substitution allows the substituting block to be sourced from within or outside the plan. When the substituting block is from within the plan, we refer to the substitution as an internal block-substitution.
\subsubsection{Threats in Block-Substitution}

\begin{figure}[!bp]

\begin{subfigure}{.32\columnwidth}
     \centering
        \begin{tikzpicture}[node distance= 1.2cm]
            \node (init)[process]{INIT};
            \node (i)[process, below of= init, xshift= -.8cm]{$b_i$};
            \node (t)[process, below of= i, xshift= 1.6cm, yshift=.3cm]{$\hat{b}_x$};
            \node (j)[process, below of= t, xshift= -1.6cm, , yshift=.3cm]{$b_j$};
            \node (goal)[process, below of= j,xshift = .8cm ]{GOAL};
           
            \draw [arrow, dotted](init) -> (i);
            \draw [arrow, dotted](init) -> (t);
            \draw [arrow, dotted](i) -> (t);
            \draw [arrow, dotted](t) -> (j);
            \draw [arrow](i) -> node[left, xshift=.1cm]{$PC(\langle v,d \rangle)$}(j);
             \draw [arrow, dotted](j) -> (goal);
             \draw [arrow, dotted](t) -> (goal);
        \end{tikzpicture}
        \caption{}
        \label{fig:threat-1a}
\end{subfigure}
\begin{subfigure}{.32\columnwidth}
     \centering
        \begin{tikzpicture}[node distance= 1.2cm]
            \node (init)[process]{INIT};
            \node (i)[process, below of= init, xshift= -.8cm]{$b_i$};
            \node (t)[process, below of= i, xshift= 1.6cm, yshift=.3cm]{$\hat{b}_x$};
            \node (j)[process, below of= t, xshift= -1.6cm, , yshift=.3cm]{$b_j$};
            \node (goal)[process, below of= j,xshift = .8cm ]{GOAL};
           
            \draw [arrow, dotted](init) -> (i);
            \draw [arrow, dotted](init) -> (t);
            \draw [arrow, dotted](i) -> (t);
            \draw [arrow, dotted, thick](t) -> (j);
            \draw [out=-20, in=-20, arrow, thick](j) to node[right, xshift=-.5cm, yshift=-.3cm]{$CD(\langle v,d \rangle)$}(t);
            \draw [arrow](i) -> node[left, xshift=.1cm]{$PC(\langle v,d \rangle)$}(j);
             \draw [arrow, dotted](j) -> (goal);
             \draw [arrow, dotted](t) -> (goal);
        \end{tikzpicture}
        \caption{}
        \label{fig:threat-1b}
\end{subfigure}
\begin{subfigure}{.32\columnwidth}
     \centering
        \begin{tikzpicture}[node distance= 1.2cm]
            \node (init)[process]{INIT};
            \node (i)[process, below of= init, xshift= -.8cm]{$b_i$};
            \node (t)[process, below of= i, xshift= 1.6cm, yshift=.3cm]{$\hat{b}_x$};
            \node (j)[process, below of= t, xshift= -1.6cm, , yshift=.3cm]{$b_j$};
            \node (goal)[process, below of= j,xshift = .8cm ]{GOAL};
           
            \draw [arrow, dotted](init) -> (i);
            \draw [arrow, dotted](init) -> (t);
            \draw [arrow, dotted](i) -> (t);
            \draw [arrow, dotted, thick](t) -> (j);
            \draw [out=20, in=20, arrow, thick](t) to node[right, xshift=-.5cm, yshift=.3cm]{$DP(\langle v,d \rangle)$}(i);
            \draw [arrow](i) -> node[left,xshift=.1cm]{$PC(\langle v,d \rangle)$}(j);
             \draw [arrow, dotted](j) -> (goal);
             \draw [arrow, dotted](t) -> (goal);
        \end{tikzpicture}
        \caption{}
        \label{fig:threat-1c}
\end{subfigure}
\caption{
Formation of cycles in a BDPO plan due to a promotion or demotion ordering, where (a) a block $\hat{b}_x$ with $\langle v, d \rangle \in del(\hat{b}_x)$ threatens a causal link $b_i \xrightarrow{\langle v, d \rangle} b_j$, and $b_i\prec \hat{b}_x \prec b_j$. To resolve this threat, (b) adding $CD(\langle v, d \rangle) \in Re(b_j \prec \hat{b}_x)$ leads to a cycle $b_j \prec \hat{b}_x \prec b_j$, while (c) adding $DP(\langle v, d \rangle) \in Re(\hat{b}_x \prec b_i)$ also induce a cycle $b_i \prec \hat{b}_x \prec b_i$, both resulting in a invalid plan.
}
\label{fig:threat-1}
\end{figure}

It is crucial to analyze scenarios where a block-substitution may introduce threats, and to develop effective threat-resolving strategies. Let $\pi_{bdp}= \langle \mathcal{O},\mathcal{B}, \prec\rangle$ be a valid BDPO plan, and $b_x\in \mathcal{B}$ be a block. A block-substitution of  $b_x$ with $\hat{b}_x$ produces a BDPO plan $\pi'_{bdp}= \langle \mathcal{O}', \mathcal{B}', \prec'\rangle$, where threats can arise in two scenarios; 1) $\hat{b}_x$ poses a threat to a causal link $b_i \xrightarrow{\langle v, d \rangle}b_j$ where $b_i, b_j \in \mathcal{B}'$, and 2) a block $b_t$ becomes a threat to  $b_k \xrightarrow{\langle v, d\rangle} \hat{b}_x$ where $b_t, b_k \in \mathcal{B}'$.

Let us investigate the first scenario where the substituting block $\hat{b}_x$ can threaten other causal links in  $\pi'_{bdp}$. Let $b_i \xrightarrow{\langle v, d\rangle} b_j$ be a causal link in $\pi'_{bdp}$ and $\langle v, d\rangle \in del(\hat{b}_x)$. Block $\hat{b}_x$ threatens the causal link if it is not ordered before $b_i$ or after $b_j$.  When $\hat{b}_x$ threatens  $b_i \xrightarrow{\langle v, d\rangle} b_j$,  this threat can be resolved by adding  $DP(\langle v, d\rangle)$ to $Re(\hat{b}_x \prec b_i)$ or $CD(\langle v, d\rangle)$ to $Re(b_j \prec \hat{b}_x)$, except when the additional ordering introduces cycle in $\pi'_{bdp}$. Adding CD or DP reasons to resolve threats can introduce cycle in two situations; Situation (1): $\hat{b}_x$ is ordered between $b_i$ and $b_j$, i.e., $b_i\prec \hat{b}_x \prec b_j$ (illustrated in Figure \ref{fig:threat-1}), and situation (2): block $b_i$ provides $\langle v, d \rangle$ to both $b_j$ and $\hat{b}_x$,  where $\langle v, d \rangle \in (del(b_j) \cap del(\hat{b}_x))$. In latter situation (demonstrated in Figure \ref{fig:threat-2}), $\hat{b}_x$ threatens $b_i \xrightarrow{\langle v, d \rangle} b_j$, and $b_j$ threatens
$b_i \xrightarrow{\langle v, d \rangle} \hat{b}_x$ as well.   Adding promotion or demotion orderings to resolve these threats in both situations invalidates the plan by inducing a cycle. 
These threats can be resolved by substituting $b_j$ with $\hat{b}_x$ only when the block $b_j$ becomes redundant in  $\pi'_{bdp}$. The block $b_j$ is identified as redundant, if $\hat{b}_x$  produces all the facts that $b_j$ provides to other blocks through causal links, i.e.,  $\forall b_j \xrightarrow{\langle v, d \rangle} b_k$,  $\langle v, d\rangle\in prod(\hat{b}_x)$ s.t. $b_k\in \mathcal{B}'$.  

Let us now consider second scenario in which a block $b_t$ (where $b_t \ne \hat{b}_x$) in $\pi'_{bdp}$  poses a threat to a causal link  $b_k \xrightarrow{\langle v, d \rangle} \hat{b}_x$ where $b_k\in (\mathcal{B}'-\{b_t, \hat{b}_x\})$. Similar to the first scenario, promotion or demotion orderings can not resolve this threat if $b_t$ is ordered between $b_k$ and $\hat{b}_x$. This threat can be resolved by substituting $b_t$ with $\hat{b}_x$ if $b_t$ becomes redundant in $\pi'_{bdp}$.

\begin{figure}[!bt]
    \centering
    \begin{subfigure}{.48\columnwidth}
         \centering
            \begin{tikzpicture}[node distance= 1cm]
                \node (init)[process]{INIT};
                \node (i)[process, below of= init]{$b_i$};
                
                \node (j)[process, below of= i, xshift= -1.5cm,yshift= -.5cm]{$b_j$};
                \node (k)[process, below of= i, xshift= 1.5cm, yshift= -.5cm]{$\hat{b}_x$};
                \node (goal)[process, below of= j,xshift = 1.5cm , yshift= -.6cm]{GOAL};
               
                \draw [arrow, dotted](init) -> (i);
                \draw [arrow](i) -> node[left, xshift=.1cm]{$PC(\langle v,d \rangle)$}(j);
                \draw [arrow](i) -> node[right, xshift=.1cm]{$PC(\langle v,d \rangle)$}(k);
                 \draw [arrow, dotted](j) -> (goal);
                 \draw [arrow, dotted](k) -> (goal);
            \end{tikzpicture}
            \caption{}
    \end{subfigure}
    \begin{subfigure}{.48\columnwidth}
         \centering
            \begin{tikzpicture}[node distance= 1cm]
                \node (init)[process]{INIT};
                \node (i)[process, below of= init]{$b_i$};
                
                \node (j)[process, below of= i, xshift= -1.5cm,yshift= -.5cm]{$b_j$};
                \node (k)[process, below of= i, xshift= 1.5cm, yshift= -.5cm]{$\hat{b}_x$};
                \node (goal)[process, below of= j,xshift = 1.5cm ,yshift= -.6cm]{GOAL};
               
                \draw [arrow, dotted](init) -> (i);
                \draw [arrow](i) -> node[left, xshift=.1cm]{$PC(\langle v,d \rangle)$}(j);
                \draw [arrow](i) -> node[right, xshift=.1cm]{$PC(\langle v,d \rangle)$}(k);
                 \draw [arrow, dotted](j) -> (goal);
                 \draw [arrow, dotted](k) -> (goal);
                 \draw [out=-40, in=-140, arrow, thick](j) to node[left, xshift=1cm, yshift=.4cm]{$CD(\langle v,d \rangle)$}(k);
            \end{tikzpicture}
            \caption{}
    \end{subfigure}
    \begin{subfigure}{.48\columnwidth}
         \centering
            \begin{tikzpicture}[node distance= 1cm]
                \node (init)[process]{INIT};
                \node (i)[process, below of= init]{$b_i$};
                
                \node (j)[process, below of= i, xshift= -1.5cm,yshift= -.5cm]{$b_j$};
                \node (k)[process, below of= i, xshift= 1.5cm, yshift= -.5cm]{$\hat{b}_x$};
                \node (goal)[process, below of= j,xshift = 1.5cm ,yshift= -.6cm]{GOAL};
               
                \draw [arrow, dotted](init) -> (i);
                \draw [arrow](i) -> node[left, xshift=.1cm]{$PC(\langle v,d \rangle)$}(j);
                \draw [arrow](i) -> node[right, xshift=.1cm]{$PC(\langle v,d \rangle)$}(k);
                 \draw [arrow, dotted](j) -> (goal);
                 \draw [arrow, dotted](k) -> (goal);
                 \draw [out=-40, in=-140, arrow, thick](j) to node[left, xshift=1cm, yshift=.4cm]{$CD(\langle v,d \rangle)$}(k);
                  \draw [out=140, in=40, arrow, thick](k) to node[right, xshift=-1cm, yshift=-.4cm]{$DP(\langle v,d \rangle)$}(j);
            \end{tikzpicture} 
            \caption{}
    \end{subfigure}
    \caption{Formation of cycles in a BDPO plan due to promotion and demotion orderings where, (a) block $b_i$ provides $\langle v, d\rangle$ to both $b_j$ and $\hat{b}_x$, and $\langle v, d \rangle \in (del(b_j) \cap del(\hat{b}_x))$. (b) $CD(\langle v, d \rangle)\in Re(b_j \prec \hat{b}_x )$ is added to prevent $\hat{{b}_x}$ from threatening $b_i \xrightarrow{\langle v, d \rangle} b_j$. Then, (c) $DP(\langle v, d \rangle)\in Re(\hat{b}_x \prec b_j)$ is added to prevent $b_j$ from threatening $b_i \xrightarrow{\langle v, d \rangle} \hat{b}_x$. These two additional orderings reasons lead to a cycle $b_j \prec \hat{b}_x \prec b_j$, rendering an invalid plan.}
    \label{fig:threat-2}
\end{figure}

\begin{definition} \label{resolve}
Let $\pi_{bdp}=\langle \mathcal{O},\mathcal{B}, \prec\rangle$ be a BDPO plan, and $b_t\in \mathcal{B}$ be a threat to a causal link $b_x \xrightarrow{\langle v, d \rangle} b_y$, where $\langle v, d \rangle \in del(b_t)$ and $b_x, b_y \in \mathcal{B}$. This threat can be resolved if any of the following \textbf{threat-resolving strategies} can be employed without introducing any cycle in $\pi_{bdp}$.
\begin{enumerate}
    \item \textit{Promotion}: adding an ordering $b_t \prec b_x$ to $\prec$.
    \item  \textit{Demotion}: adding an ordering $b_y \prec b_t$ to $\prec$.
    \item  \textit{Internal block-substitution}: substituting $b_y$ with $b_t$ or substituting $b_t$ with $b_y$.
\end{enumerate}
\end{definition}

During block-substitution, if any threat can not be resolved by the strategies (defined in Definition \ref{resolve}), the substitution is not valid.

\subsubsection{Block-Substitution Algorithm}
 \begin{algorithm}[!tbp]
    \caption{Substitituting a block in a block decomposed partial-order (BDPO) plan}
    \label{alg:block_substitute}
    \begin{algorithmic}[1] 
    \Statex\textbf{Input}: a BDPO plan $\pi_{bdp}=\langle \mathcal{O},\mathcal{B}, \prec\rangle$, two block $b_x\in \mathcal{B}$ and $\hat{b}_x$.
    \Statex\textbf{Output:} a BDPO plan and a boolean value 
    \Procedure{Substitute}{$\pi_{bdp}, b_x, \hat{b}_x$}
    \State $\hat{\pi}_{bdp}\equiv \langle \mathcal{O}', \mathcal{B}', \prec' \rangle\leftarrow \pi_{bdp}$
    
    \If{$\hat{b}_x\notin \mathcal{B}'$}
    \Comment{establishing causal links for $\hat{b}_x$'s precondition}
        \State $b_{new}\leftarrow \hat{b}_x$ 
        \State add $\hat{b}_x$ to $\mathcal{B}'$
        \ForAll{$\langle v, d \rangle\in \mathit{pre}(\hat{b}_x)$}
            \State find an earliest candidate producer $b$ of $\langle v, d \rangle$  for $\hat{b}_x$
            \Comment{Definition \ref{def:candidate_producer}}
            \If{$b$ is found}
             add $b \xrightarrow{\langle v, d \rangle} \hat{b}_x$ to $\prec'$
            \Else \ \Return $\pi_{bdp}$, \FALSE 
            \EndIf
        \EndFor
    \EndIf
     
        \ForAll{ $b\in \mathcal{B}'$ s.t. $b_x \xrightarrow{\langle v, d \rangle} b$} 
        \Comment{reestablishing causal links, supported by  $b_x$}
            \If{$\hat{b}_x$ produces $\langle v, d \rangle$}
            add $\hat{b}_x\xrightarrow{\langle v, d \rangle} b$ to $\prec'$
            \Else \ \Return $\pi_{bdp}$, \FALSE 
            \EndIf
        \EndFor
       \State delete $b_x$ from $\mathcal{B}'$
        
        \ForAll{threats where $b_k$ threatens $b_i\xrightarrow{\langle v, d\rangle} b_j$ s.t  $b_i, b_j, b_k\in \mathcal{B}'$}
        \Comment{resolving threats}

        \If{$b_k\nprec b_j$} $\eta \leftarrow b_j\prec b_k$
            \Comment{demotion ordering}
            \Else\ $\eta \leftarrow b_k\prec b_i$ \Comment{promotion ordering}
        \EndIf
        \If{adding $\eta$ to $\prec'$ renders no cycle in $\hat{\pi}_{bdp}$}
          add $\eta$ to $\prec'$ 
        \Else 
        \Comment{try internal substitution by $\hat{b}_x$}
           \If{$b_k = b_{new}$}
            ($\hat{\pi}_{bdp}, success$)$\leftarrow$ \Call{Substitute}{$\hat{\pi}_{bdp}$, $b_j, b_k$}
            \ElsIf{$b_j = b_{new}$}
            ($\hat{\pi}_{bdp}, success$)$\leftarrow$ \Call{Substitute}{$\hat{\pi}_{bdp}$, $b_k, b_j$}
            \Else \ \Return $\pi_{bdp}$, \FALSE 
            \EndIf
             \If{$success$ is \FALSE}  
              \Return $\pi_{bdp}$, \FALSE 
             \EndIf
        \EndIf
        \EndFor
        \State \Return $\hat{\pi}_{bdp}$, \TRUE 
    \EndProcedure
   
    \end{algorithmic}
\end{algorithm}
 The block-substitution procedure, named SUBSTITUTE, (Algorithm \ref{alg:block_substitute}) takes a valid BDPO plan $\pi_{bdp}=\langle \mathcal{O}, \mathcal{B}, \prec \rangle$ along with two blocks $b_x$ and $\hat{b}_x$ as input, where $b_x\in \mathcal{B}$, and produces a valid BDPO plan for a planning task $\Pi$. The block $\hat{b}_x$ can be from inside or outside  $\pi_{bdp}$. 

This procedure can be divided into three main parts. First, it ensures each precondition of $\hat{b}_x$ is supported by a causal link.  If it is an internal substitution (i.e., $\hat{b}_x \in \mathcal{B}$), the preconditions of $\hat{b}_x$ are already supported by causal links since $\pi_{bdp}$ is valid. When $\hat{b}_x$ is an external block, i.e.,  $\hat{b}_x \notin \mathcal{B}$, the procedure adds $\hat{b}_x$ to $\pi_{bdp}$, followed by establishing a causal link with the earliest candidate producer (Definition \ref{def:candidate_producer}) for each precondition belonging to $\hat{b}_x$ (lines 3-12). The substitution is considered unsuccessful if a causal link can not be established for a precondition of $\hat{b}_x$. Second, the procedure reinstates the causal links, previously supported by the old block $b_x$, using $\hat{b}_x$ as the new producer (lines 13-17). However, if $\hat{b}_x$ does not produce any fact $\langle v,d \rangle$ such that $b_x \xrightarrow{\langle v,d \rangle} b$ for a block $ b\in \mathcal{B}$, the substitution is deemed unsuccessful. After establishing the required causal links, the block $b_x$ is removed from the plan. Then, the procedure identifies the threats introduced during this process, and applies the threat-resolving strategies, defined in Definition \ref{resolve}, to resolve them (lines 19-32). When the procedure successfully resolves all threats, it returns the resultant BDPO plan. It is essential to highlight that the scenarios depicted in Figures \ref{fig:threat-1} and \ref{fig:threat-2} necessitate an internal block-substitution resolving strategy. To resolve the threats in these situations, we only consider replacing the conflicting block with $\hat{b}_x$, assuming that the conflicting block in the resultant BDPO plan becomes redundant only after the substitution of $b_x$ with $\hat{b}_x$.

The worst case complexity for the block-substitution algorithm (Algorithm \ref{alg:block_substitute}) is $O(n^2p^2)$, where $n$ is the number of operators in the plan, and $p$ is the maximum number of facts in a precondition or an effect of any plan operator. Establishing causal links for substituting block and those supported by the original block run at most $np$ and $p$ times, respectively. Subsequently, the threat resolving loop runs in $O(n^2p^2)$ time. Substituting the conflicting block with $\hat{b}_x$ to resolve a threat takes only $p$ time. Since this internal substitution does not require establishing preconditions of $\hat{b}_x$, and does not introduce any new threat. Therefore, worst case complexity of the algorithm can be expressed as  $O(np+p+n^2p^2)= O(n^2p^2)$. 
\begin{theorem} \label{theorem:correct_substitute}
[Correctness of Block-Substitution Algorithm] Given a valid BDPO plan $\pi_{bdp}=\langle \mathcal{O},\mathcal{B}, \prec\rangle$, a successful block-substitution of a block $b_x\in \mathcal{B}$ with a block $\hat{b}_x$ yields a valid BDPO plan.
\end{theorem}
\begin{proof}
Since $\pi_{bdp}$ is valid, every block precondition in $\pi_{bdp}$ is supported by a causal link with no threat \cite{siddiqui_patrik_2012}.

A successful block-substitution of $b_x$ with $\hat{b}_x$, results in a BDPO plan $\pi'_{bdp}=\langle \mathcal{O'},\mathcal{B'} \prec'\rangle$ where all required causal links for the substituting block $\hat{b}_x$ are established, and the causal links, supported by the original block $b_x$, are re-established using $\hat{b}_x$ as producer. Since the preconditions of $\hat{b}_x$ are supported by the predecessors of $b_x$, and the blocks, whose preconditions are now provided by $\hat{b}_x$, were previously successors of $b_x$, the algorithm introduce no cycle in $\pi'_{bdp}$ while establishing the required causal links.

We will prove that no causal link in $\pi'_{bdp}$ has a threat. Let us first consider the causal links
 $b_i \xrightarrow{\langle v,d \rangle} b_j$ in $\pi'_{bdp}$  that are preexisted in $\pi_{bdp}$, i.e., $b_i, b_j \in \mathcal{B} \cap \mathcal{B}'$. No block $b\neq \hat{b}_x$ threatens $b_i \xrightarrow{\langle v,d \rangle} b_j$ in $\pi'_{bdp}$. If such a block $b$ were to threaten this causal link, it would also threaten $b_i \xrightarrow{\langle v,d \rangle} b_j$ in $\pi_{bdp}$. This cannot be true since $\pi_{bdp}$ is valid.  The situations where the substituting block $\hat{b}_x$ can pose a threat to  $b_i \xrightarrow{\langle v,d \rangle} b_j$ are thoroughly examined in Section  3.1.1 and illustrated in Figures \ref{fig:threat-1} and \ref{fig:threat-2}.  As the block-substitution is successful, those threats are resolved using the threat-resolving strategies defined in Definition \ref{resolve}.  
 
 Now, let us consider the causal links that are associated with $\hat{b}_x$. The causal links of the form $\hat{b}_x \xrightarrow{\langle v, d \rangle} b_j$ in $\pi'_{bdp}$ do not have any threat. If such a threat were present, the corresponding causal link $b_x \xrightarrow{\langle v, d \rangle} b_j$ in the original plan $\pi_{bdp}$ would also be threatened. This contradicts our initial assumption as $\pi_{bdp}$ is valid. Lastly, all threats to the causal links of the form $b_i \xrightarrow{\langle v, d \rangle} \hat{b}_x$ in $\pi'_{bdp}$ are resolved  (as discussed in Section 3.1.1) since the block-substitution is successful. Notably, the algorithm checks for cycles after adding each threat resolving ordering. If a cycle is introduced, the algorithm attempts to resolve it by internal block-substitution, and if it fails, the algorithm regards the block-substitution process as unsuccessful. Hence any successful substitution yields an acyclic valid BDPO plan.
\end{proof}

Having established the correctness of Algorithm~\ref{alg:block_substitute}, we now examine its completeness. Notably, the Block-Substitution Algorithm always selects the earliest producer to establish causal links for the preconditions of the substituting block. Committing to the earliest candidate producer not only reduces the chance of unnecessary transitive ordering but also avoids exploring all possible producer combinations for the preconditions of the substituting block. However, this choice comes at the cost of missing some valid solutions in particular scenarios. The following theorem and proof identify the exact conditions under which the algorithm cannot provide a solution by substitution, though a valid substitution exists.

\begin{theorem}\label{theorem:incomplete_substitute}
[Incompleteness of Block-Substitution Algorithm] Algorithm \ref{alg:block_substitute} is not complete. That is, there exists a valid BDPO plan $\pi_{bdp}=\langle \mathcal{O},\mathcal{B}, \prec\rangle$ and blocks $b_x \in \mathcal{B}$ and $\hat{b}_x$, such that a valid BDPO plan $\pi'_{bdp}$ can be obtained by substituting $b_x$ with $\hat{b}_x$, but Algorithm 2 does not provide $\pi'_{bdp}$.
\end{theorem}
\begin{proof}
We will prove by constructing a specific BDPO plan and substitution scenario where a valid substitution exists but Algorithm \ref{alg:block_substitute} fails to find it.

Consider a BDPO plan $\pi_{bdp} = \langle \mathcal{O}, \mathcal{B}, \prec \rangle$ with blocks $\mathcal{B} = \{b_1, b_2, b_t, b_x\}$ and ordering constraints $b_1 \prec b_2 \prec b_x$ and $b_1\prec b_t$. Both $b_1$ and $b_2$ produce $\langle v,d \rangle$ and there exists a causal link $b_1 \xrightarrow{\langle v,d \rangle} b_t$. Let $b_x$ be the original block and $\hat{b}_x$ be the substituting block with preconditions $\mathit{pre}(\hat{b}_x) = \{\langle v,d \rangle\}$ where $\langle v,d \rangle \notin \mathit{pre}(b_x)$. Also, blocks $b_t$ and $\hat{b}_x$ both delete the fact $\langle v,d \rangle$

When substituting $b_x$ with $\hat{b}_x$, Algorithm~\ref{alg:block_substitute} selects $b_1$ to construct causal link $b_1 \xrightarrow{\langle v,d \rangle} \hat{b}_x$ to provide $\langle v,d \rangle$ to $\hat{b}_x$ since $b_1$ is the earliest producer. However, this creates a mutual threat between blocks $b_t$ and $\hat{b}_x$ since they both consume $\langle v,d \rangle$ from $b_1$ and threaten each other's causal links. Resolving these bidirectional threats requires contradictory ordering constraints that introduce a cycle, as illustrated in Figure~\ref{fig:threat-2}. Therefore, the algorithm returns false.

However, a valid substitution exists by binding the precondition $\langle v,d \rangle$ of $\hat{b}_x$ with producer $b_2$ instead of $b_1$ and creating an ordering $b_t \prec b_2$. This will avoid the mutual threat scenario, yielding a valid BDPO plan. Since Algorithm~\ref{alg:block_substitute} commits to the earliest producer $b_1$ and fails to explore this alternative, it is incomplete.

\end{proof}

\subsection{Flexibility Improvement via Block-Substitution (FIBS) Algorithm}\label{subsection:fibs}
Unlike EOG and block deordering, our proposed algorithm enhances plan flexibility by altering a plan strategically using block substitution. We refer to this algorithm as the Flexibility Improvement via Block-Substitution (FIBS) algorithm.

The core of the FIBS algorithm is the RESOLVE procedure (Algorithm \ref{alg:resolve}), which eliminates ordering constraints between pairs of blocks through block-substitution. RESOLVE procedure takes a BDPO plan $\pi_{bdp}=\langle \mathcal{O},\mathcal{B}, \prec\rangle$ for a planning task $\Pi$ and  two blocks $b_i, b_j\in \mathcal{B}$ as input, where there is a basic ordering, $b_i \prec b_j$. It seeks a suitable candidate block  $\hat{b}_j$ to replace  $b_j$ so that there will be no ordering between $b_i$ and $\hat{b}_j$, essentially improving the $\mathit{flex}$ of $\pi_{bdp}$. Every block in a BDPO plan is a partial-order subplan. To find suitable subplans for substituting block $b_j$, we design a subtask $\Pi_{sub}$ corresponding to $b_j$. 
 
\begin{algorithm}[!t]
    \caption{Resolve Ordering between a pair of blocks via block-substitution}
    \label{alg:resolve}
    \begin{algorithmic}[1] 
    \Statex\textbf{Input}:  A valid BDPO plan $\pi_{bdp}=\langle \mathcal{O},\mathcal{B} \prec\rangle$ for a planning task $\Pi= \langle \mathcal{V}, O, s_0, s_*\rangle$, and two blocks $b_i, b_j \in \mathcal{B}$ s.t. there is a basic ordering  $b_i \prec b_j$.
    \Statex\textbf{Output:} A BDPO plan, and a boolean value 
    \Procedure{Resolve}{$\Pi, \pi_{bdp}, b_i, b_j$}
        \State  $\hat{\pi}\leftarrow$ linearize operators in blocks $b\in \mathcal{B}$ s.t.
        $b \neq b_i$, $b_I \prec b \prec b_j$, and $b_I=\{o_I\}$ 
        \State $\hat{s}_0 \leftarrow$ $apply(\hat{\pi},s_0)$ \Comment{get initial state for subtask}
        \State $G \leftarrow \{\langle v,d \rangle \mid b_j \xrightarrow{\langle v,d \rangle} b_k, b_k \in \mathcal{B}$\}
         \State $C \leftarrow \{\langle v,d \rangle \mid b_x \xrightarrow{\langle v,d \rangle} b_y$, $b_x \prec b_j \prec b_y, b_x \ne b_i,\{b_x, b_y\}\in \mathcal{B}\}$ 
          \State  $\hat{s}_* \leftarrow G \cup C$ \Comment{get goal for subtask}
        \State construct a subtask $\Pi_{sub}=\langle \mathcal{V}, O,\hat{s}_0$, $\hat{s}_*\rangle$ 
            \State $subplans$ $\leftarrow$  \Call {generate\_plans}{$\Pi_{sub} $} 
            \ForAll{$\hat{\pi} \in subplans$}
                \State $\hat{\pi}_{pop} \leftarrow$ apply EOG on $\hat{\pi}$
                \State make a new block $\hat{b}_j$ from $\hat{\pi}_{pop}$ 
                \State ($\hat{\pi}_{bdp}, success$) $\leftarrow$  \Call{Substitute}{$\pi_{bdp}$, $b_j, \hat{b}_j$}
                \If{ $success$ \textbf{and} $\mathit{flex}(\hat{\pi}_{bdp}) > \mathit{flex}(\pi_{bdp})$ \textbf{and} $cost(\hat{\pi}_{bdp}) \leq cost(\pi_{bdp})$}
                    \State \Return $\hat{\pi}_{bdp}$, \TRUE
                \EndIf
            \EndFor
        \State\Return  $\pi_{bdp}$, \FALSE
    \EndProcedure
    \end{algorithmic}
\end{algorithm}

 When establishing the initial state $\hat{s}_0$ and the goal $\hat{s}_*$ for  $\Pi_{sub}$, we assume that the predecessors and successors (except $b_i$) of the candidate substituting block $\hat{b}_j$ will be the same as $b_j$. We construct the initial state by progressing the operators in blocks (except $b_i$) preceding $b_j$ from the initial state $s_0$ of the planning task $\Pi$. This has been achieved by creating a subplan $\pi'$ through linearization of block operators (except $b_i$) before $b_j$, and then, $\hat{s}_0$ is set to the state produced by applying $\pi'$  in $s_0$. When the ordering is $b_i \prec b_j$, $b_i$ precedes $b_j$. We exclude the operators of $b_i$ from $\pi'$ to ensure that operators or subplans applicable in the state $\hat{s}_0$ do not require any fact from $b_i$. Dropping $b_i$'s operators from $\pi'$ does not invalidate  $\pi'$, because there is no other block $b'$ in $\pi'$ with $b_i \prec b'$  since $b_i \prec b_j$ is a basic ordering.
 
 We estimate the goal for $\Pi_{sub}$ by aggregating the facts (denoted as $G$) that $b_j$ achieves for other blocks, and the facts (denoted as $C$) that $b_j$'s predecessors (except $b_i$) achieve for its successors in $\pi_{bdp}$. It is important to note that facts in $C$ also belong to $\hat{s}_0$, i.e., $C\subset \hat{s}_0$. Therefore, the solutions of $\Pi_{sub}$ do not need to produce the facts in $C$ but can not delete them. We include these facts in the goal so that the candidate block $\hat{b}_j$ does not delete the facts that its predecessor provides to its successor. Because in that case, $\hat{b}_j$ will pose threats to those causal links, and adding promotion or demotion ordering to resolve these threats will produce a cycle in the resultant BDPO plan (illustrated in Figure \ref{fig:threat-1}).
 
We employ an off-the-shelf cost-bounded planner (c.g., the LAMA planner by \citeauthor{lama} \citeyear{lama}) to generate multiple solutions $\hat{\pi}$ (i.e., subplans) for $\Pi_{sub}$ with a time-bound. We set the cost bound to the cost of $b_j$. By applying EOG, this procedure converts $\hat{\pi}$ into a partial-order subplan $\hat{\pi}_{pop}$, and then creates a candidate block $\hat{b}_j$  from $\hat{\pi}_{pop}$.  When substituting $b_j$ with $\hat{b}_j$ is successful, the RESOLVE procedure accepts the resultant BDPO plan $\hat{\pi}_{bdp}$ if $\hat{\pi}_{bdp}$ satisfies \emph{Relative Flexibility Optimization} (RFO) criteria (Definition \ref{RFO}) w.r.t. $\pi_{bdp}$. This criterion ensures that plan flexibility and cost are not compromised in this procedure.

\begin{definition}\label{RFO}
    Let $\pi$ and $\pi'$ be two valid POPs for a planning task $\Pi$.  The POP $\pi'$ is a \textbf{relative flexibility optimization (RFO)} of $\pi$ if $\mathit{flex}(\pi') > \mathit{flex}(\pi)$, and $cost(\pi') \le cost(\pi)$
\end{definition}

The FIBS algorithm employs the \textsc{resolve} procedure for every basic ordering in a BDPO plan to enhance its flexibility iteratively on top of EOG and block deordering, while maintaining the plan cost under the RFO criteria. 

\begin{algorithm}[!tb]
    \caption{Flexibility Improvement via Block-Substitution (FIBS)}
    \label{alg:fibs}
    
    \begin{algorithmic}[1] 
    \Statex\textbf{Input}: A valid sequential plan $\pi$ for a planning task $\Pi$.
    \Statex\textbf{Output:} A valid BDPO plan
    \State $\pi_{pop} \equiv \langle \mathcal{O}, \prec \rangle \leftarrow$ \Call{EOG}{$\pi$}
    \Comment{generate POP using EOG}
    \State build $\pi_{bdp}= \langle \mathcal{O}, \mathcal{B}, \prec \rangle$, where for every $o\in \mathcal{O}$, there is a block $b=\{o\}$ in $\mathcal{B}$
    \State $\pi_{sd1}\leftarrow$ \Call{Substitution-Deorder}{$\Pi,\pi_{bdp}$} \Comment{substitute only primitive blocks}
    \State $\pi_{bd}\leftarrow$ \Call{Block-Deorder} {$\pi_{sd1}$} 
    \Comment{build compound blocks}
    \State $\pi_{sd2}\leftarrow$ \Call{Substitution-Deorder}{$\Pi, \pi_{bd}$} \Comment{substitute both primitive and compound blocks}
    \State \Return $\pi_{sd2}$
    \Procedure{Substitution-Deorder}{$\Pi, \pi_{bdp}$}
        \ForAll{basic ordering $ (b_i \prec b_j) \in \prec$}
            \State ($\pi_{bdp}, success$) $\leftarrow$ \Call{Resolve}{$\Pi, \pi_{bdp}, b_i, b_j$}
            \Comment{try substituting $b_j$}
            \If{$success$ is \FALSE}
                \State ($\pi_{bdp}, success$) $\leftarrow$ \Call{Resolve}{$\Pi, \pi_{bdp}, b_j, b_i$}
                \Comment{try substituting $b_i$}
            \EndIf
            \If{$success$} 
            \State \Return \Call{Substitution-Deorder}{$\Pi, \pi_{bdp}$}
            
            \EndIf
        \EndFor
        \State \Return $\pi_{bdp}$
    \EndProcedure
    \end{algorithmic}
\end{algorithm}

The FIBS algorithm  (presented in Algorithm \ref{alg:fibs}) takes a valid sequential plan  $\pi$ as input for a planning task $\Pi$, and generates a valid BDPO plan for $\Pi$. FIBS enhances the flexibility of $\pi$ in four phases. The first phase converts $\pi$ into a partial-order plan $\pi_{pop}= (\mathcal{O}, \prec)$ using EOG.  $\pi_{pop}$ is then transformed into a BDPO plan $\pi_{bdp} = (\mathcal{O}, \mathcal{B}, \prec)$ by adding a block $b=\{o\}$ for each operator $o\in \mathcal{O}$. Following this, SUBSTITUTION-DEORDER (\emph{SD}) procedure (lines 7-17) is applied to $\pi_{bdp}$ to eliminate orderings by substituting blocks. The SUBSTITUTION-DEORDER procedure systematically examines each basic ordering constraint $b_i \prec b_j$ and attempts to eliminate it through substitution.  In this phase, referred to as \emph{SD1}, only primitive blocks are substituted as no compound block is formed yet. The next phase, block deodering (\emph{BD}), encapsulates coherent primary blocks in compound blocks to eliminate further orderings in the BDPO plan. The final phase, termed \emph{SD2}, employs the SUBSTITUTION-DEORDER procedure again to replace both primitive and compound blocks to minimize operator orderings. Therefore, we perform SUBSTITUTION-DEORDER twice, once before block deordering and once afterward, to distinguish and evaluate the performance of primitive and compound block-substitution. 

The SUBSTITUTION-DEORDER procedure (lines 7-17) in Algorithm \ref{alg:fibs} takes each basic ordering from the beginning of a BDPO plan, and calls the RESOLVE procedure (Algorithm \ref{alg:resolve}) to eliminate the ordering by replacing blocks. SUBSTITUTION-DEORDER attempts to remove an ordering $b_i\prec b_j$ first by substituting $b_j$, and if that fails, it tries to remove the ordering by replacing $b_i$. After each successful ordering removal, SUBSTITUTION-DEORDER restarts from the beginning of the latest BDPO plan. This procedure ends when there is no successful ordering removal in a complete examination of all orderings.

\begin{theorem}[Correctness of Resolve Algorithm] \label{theorem:correctness_resolve}
Given a valid BDPO plan $\pi_{bdp} = \langle \mathcal{O}, \mathcal{B}, \prec \rangle$ for a planning task $\Pi$ and two blocks $b_i, b_j \in \mathcal{B}$ such that $b_i \prec b_j$, Algorithm~\ref{alg:resolve} always produces a valid BDPO plan for $\Pi$. When the algorithm succeeds, it yields an RFO of $\pi_{bdp}$.
\end{theorem}
\begin{proof}
We prove that Algorithm~\ref{alg:resolve} always returns a valid BDPO plan for the given planning task, and that whenever it succeeds, the resulting plan is an RFO of the given BDPO plan. 

The algorithm is considered \emph{successful} if it accepts a BDPO plan produced by block substitution; otherwise, it is regarded as \emph{unsuccessful}. In the unsuccessful scenario, the algorithm simply returns the original plan $\pi_{bdp}$, which is valid by assumption. In the successful case, the algorithm returns a plan $\pi'_{bdp}$ obtained by block substitution. By Theorem~\ref{theorem:correct_substitute}, any successful block substitution produces a valid BDPO plan. Furthermore, Algorithm~\ref{alg:resolve} only accepts a substitution when both the flexibility and cost conditions are satisfied, $\mathit{flex}(\pi'_{bdp}) > \mathit{flex}(\pi_{bdp})$ and $\mathit{cost}(\pi'_{bdp}) \leq \mathit{cost}(\pi_{bdp})$, ensuring that $\pi'_{bdp}$ is an RFO of $\pi_{bdp}$.

Finally, Algorithm~\ref{alg:resolve} guarantees that the resulting BDPO plan is valid with respect to the planning task $\Pi$. When constructing the subtask $\Pi_{sub}$ to generate candidate substitutions for $b_j$, the algorithm defines the goal of $\Pi_{sub}$ as the set $G$ of all facts provided by $b_j$ that are required by its successors or by the overall goal of $\Pi$. Thus, whenever a valid subplan for $\Pi_{sub}$ is found and successfully substitutes $b_j$, the resulting BDPO plan preserves all necessary causal links and remains valid for $\Pi$.
\end{proof}

\begin{theorem}[Incompleteness of Resolve Algorithm]\label{theorem_incomplete_resolve}
The Algorithm~\ref{alg:resolve} is not complete. That is, there exist valid BDPO plan $\pi_{bdp} = \langle \mathcal{O}, \mathcal{B}, \prec \rangle$ for a planning task $\Pi$ and blocks $b_i, b_j \in \mathcal{B}$ with $b_i\prec b_j$ such that a valid BDPO plan $\pi'_{bdp}$, where $\pi'_{bdp}$ is an RFO of  $\pi_{bdp}$, for $\Pi$ can be constructed by substituting $b_j$, but Algorithm~\ref{alg:resolve} does not provide $\pi'_{bdp}$.
\end{theorem}
\begin{proof}
The incompleteness of Algorithm~\ref{alg:resolve} stems from three sources:
(1) When constructing the subtask to generate alternative subplans (i.e., blocks) to substitute block $b_j$, the algorithm assumes that the predecessors and successors of the new substituting block will coincide with those of $b_j$. This assumption prevents the exploration of subplans that require establishing causal links with blocks currently unordered with $b_j$. However, there may exist valid solutions, constructed by substituting $b_j$ with such a subplan, but still yield better overall flexibility because the substitution eliminates more existing orderings than it introduces. 
(2) The completeness of this algorithm is limited by the underlying planner used in the \textsc{Generate-Plans} procedure, which may not produce all possible valid subplans for $\Pi_{sub}$; Lastly,
(3) The incompleteness of this algorithm follows the incompleteness from the \textsc{Substitute} procedure (Theorem~\ref{theorem:incomplete_substitute}).

\end{proof}
\begin{theorem}[Correctness of FIBS Algorithm]
Given a valid sequential plan $\pi$ for a planning task $\Pi$, Algorithm~\ref{alg:fibs} always yields a valid BDPO plan for $\Pi$.
\end{theorem}
\begin{proof}
    Algorithm~\ref{alg:fibs} constructs POP and then a BDPO plan from a given sequential plan for a planning task using EOG and Block Deordering, respectively. The correctness of EOG and Block Deordering established in the literature \cite{kk,siddiqui_patrik_2012}. Since Algorithm \ref{alg:fibs} modifies the BDPO plan, generated by block deordering, solely through the \textsc{Resolve} (Algorithm \ref{alg:resolve}) procedure. The correctness of Algorithm~\ref{alg:fibs} follows directly from the correctness of Algorithm \ref{alg:resolve} established in Theorem \ref{theorem:correctness_resolve}.
\end{proof}

\begin{theorem}[Incompleteness of FIBS Algorithm]
Algorithm~\ref{alg:fibs} is not complete. That is, there exists a valid sequential plan $\pi$ for a planning task $\Pi$ such that a valid BDPO plan $\pi'_{bdp}$  for  $\Pi$ exists, where $\pi'_{bdp}$ is an RFO of the resultant BDPO plan $\pi_{bdp}$ of the Algorithm~\ref{alg:fibs}.
\end{theorem}
\begin{proof}
    Algorithm \ref{alg:fibs} improves the BDPO plan, generated from the given sequential plan, through the \textsc{Resolve} (Algorithm \ref{alg:resolve}) procedure. Since Algorithm \ref{alg:resolve} is incomplete (Theorem~\ref{theorem_incomplete_resolve}), \textsc{FIBS} (Algorithm \ref{alg:fibs}) inherits the incompleteness as well.
\end{proof}
\subsection{Removing Redundant Operators in FIBS}\label{section:fibs_pruning}
We also incorporate plan reduction within the FIBS algorithm by applying either backward justification or greedy justification to BDPO plans. Inherently, block deordering captures inverse operators (Definition \ref{def:inverse_operator}) by enclosing them in a block, rendering the block redundant. Besides inverse operators, redundant blocks can also be identified using backward justification and greedy justification. Following \citeauthor{Fink_yang1997} (\citeyear{Fink_yang1997}), we define \emph{Backward Justification} and \emph{Greedy Justification} with respect to a BDPO plan.

\begin{definition}
    Let $\pi_{bdp}= \langle \mathcal{O}, \mathcal{B}, \prec \rangle$ be a valid BDPO plan for a planning task $\Pi = \langle \mathcal{V}, \mathcal{O}, s_0, s_* \rangle$. Then:
    \begin{itemize}
        \item A block $b \in \mathcal{B}$ is called \textbf{backward justified} if it produces some fact necessary for achieving the goal state $s_*$. 
        \item A block $b \in \mathcal{B}$ is called \textbf{greedy justified} if removing $b$ and all subsequent blocks of $b$ that depend on it renders the resultant BDPO plan invalid. 
    \end{itemize}
\end{definition}

A block in $\pi_{bdp}$ can therefore be backward justified either by producing a fact $\langle v,d \rangle$ that directly contributes to the goal $s_*$, or by supporting another backward justified block in $\pi_{bdp}$. Moreover, an internal block (i.e., one enclosed within another block) may also be backward justified if it produces facts that prevent its parent block from becoming a threat to some causal link. For greedy justification, a block $b'$ is said to depend on a block $b$ if there exists a causal link $b \xrightarrow{\langle v, d \rangle} b'$ or if there exists another block $b''$ such that $b \xrightarrow{\langle v, d \rangle} b''$ and $b'$ depends (transitively) on $b''$.

After FIBS, we identify and eliminate redundant blocks from the BDPO plan using backward or greedy justification. Additionally, we often eliminate blocks during block-substitution phases (SD1 and SD2 phases) in FIBS, as a block $b$ in a BDPO plan can become redundant after a substitution is made. This happens when the substituting block can substitute $b$ without invalidating the plan. Formally, \emph{we consider a block $b \in \mathcal{B}$ as \textbf{redundant} if it is neither backward nor greedy justified, or if another block $b' \in \mathcal{B}$ can substitute $b$ without invalidating $\pi_{bdp}$}. 

To better comprehend the performance of the plan reduction strategies, we use the following relative cost optimization (RCO) criteria instead of relative flexibility optimization (RFO) criteria (Definition \ref{RFO}) in the RESOLVE procedure (Algorithm \ref{alg:resolve}). Under RFO criteria, FIBS does not compromise plan cost or flexibility. However, under RCO criteria, FIBS prioritizes plan cost over flexibility, thereby compromising flexibility for  cost reduction.

\textbf{Relative Cost Optimization (RCO) Criteria}: Let $\pi$ and $\pi'$ be two valid POPs for a planning task $\Pi$. The POP  $\pi'$ satisfies the relative cost optimization criteria w.r.t. $\pi$ if  the following condition holds :
\begin{equation*}
  cost(\pi') < cost(\pi)  \lor (\mathit{flex}(\pi') > \mathit{flex}(\pi) \land cost(\pi') = cost(\pi))
\end{equation*}

Both RCO and minimum cost least commitment POP (MCLCP) (Definition \ref{def:mclcp}) \cite{maxsat} prioritize cost over flexibility.  MCLCP is a global optimization criterion that ensures the minimum cost and maximum flexibility among all minimum-cost plans. In contrast, RCO is a local, greedy criterion that accepts any plan that either offers a strictly lower cost or maintains the same cost while improving flexibility. Thus, RCO can be seen as a computationally efficient approximation of MCLCP.

\section{Experimental Result and Discussion}

\begin{figure}[!tbp]
     \centering
     \includegraphics[width=.8\linewidth]{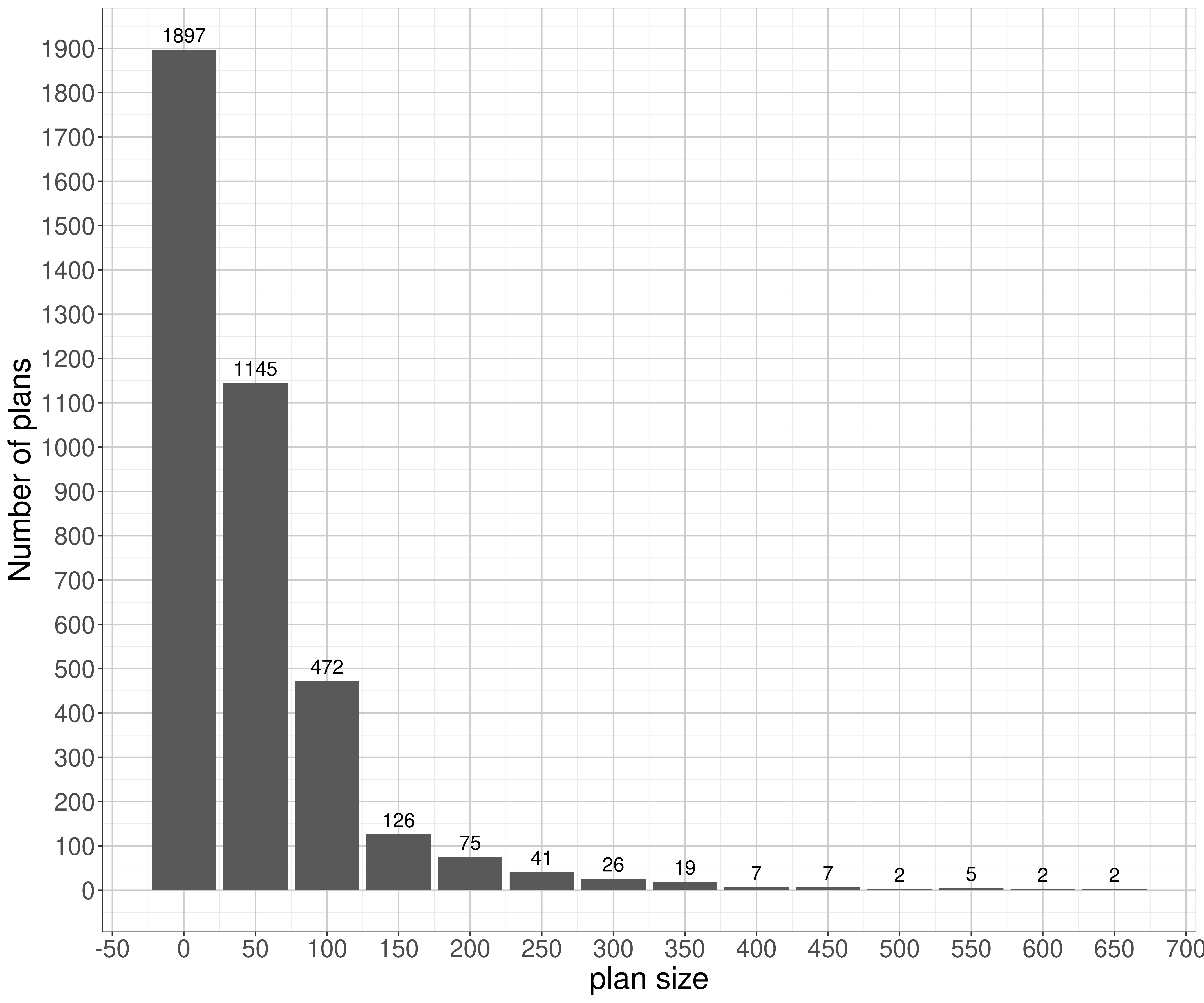}
     \caption{Histogram of plan sizes for the plans in the experimental dataset presented in Table \ref{tbl:dataset}.}
     \label{fig:histogram}
 \end{figure}

We have evaluated the FIBS algorithm (Algorithm \ref{alg:fibs}) on domains from sequential satisfying tracks of the international planning competitions (IPC). We exclude domains with conditional effects \cite{Nebel_2000}, since the algorithms presented in this paper do not consider conditional effects. For generating plans, we employed LAMA planner \cite{lama}, a two-time champion in international planning competitions, with a 30-minute time limit for each problem. LAMA is a forward search-based classical planning system that uses a landmark-based heuristic. Landmarks  \cite{landmark_Richter_2008} are propositional formulas that must hold in all possible solutions of a planning task. We have included all plans produced by LAMA within the time-bound. Table \ref{tbl:dataset} presents our dataset comprising 3826 plans from 950 problems across 32 distinct domains. In addition to the count of problems and plans, the table also provides the arithmetic average number of plan sizes, ground operators, variables, and independent variables in the planning tasks for each domain. The histogram, presented in Figure \ref{fig:histogram}, illustrates the distribution of plan sizes based on the number of plans in each size range. Most plans are relatively small, with the highest number found in the 0–50 size range, totaling 1,897 plans. This is followed by 1,145 plans in the 50–100 size range. As the plan size increases, the frequency of plans decreases sharply, with only a few exceeding a size of 300. This pattern indicates a right-skewed distribution, where larger plans are few and the majority are concentrated in smaller size categories.

\begin{table}[!tbp]
    \caption{Experimental dataset. \emph{Problems} and \emph{Plans} columns presents the number of problems and plans. The arithmetic means of the numbers of plan operators, ground operators, and variables are presented in \emph{Plan Size},  \emph{Ground Operators}, and \emph{Variables} columns, respectively. Column \emph{Plan Cost} shows the geometric mean cost of the plans. }
    \label{tbl:dataset}
    \centering
    \begin{filecontents*}{media/domain_info1.csv}
domain,problems,plans,initialsize,initialcost,operators,variables
barman,20,54,202,309,2078,310
blocks,87,234,91,91,677,38
child-snack,6,8,67,67,3183,70
depots,21,73,49,49,1247,41
elevator,150,742,58,72,881,27
floor-tile,7,44,46,81,213,21
freecell,55,208,50,50,11726,67
genome-edit-distances,20,93,114,34,7321,46
grid,5,15,53,53,7526,24
gripper,20,20,59,59,160,23
hiking,20,111,48,48,6635,14
logistics,50,161,69,69,1441,27
mystery,19,28,10,10,3522,34
mystery-prime,30,61,10,10,7492,35
no-mystery,11,28,31,29,3715,11
parc-printer,30,116,41,1156164,237,57
parking,20,144,76,74,60194,106
pathways,22,36,111,111,1031,198
peg-solitaire,30,258,30,11,185,34
pipesworld,43,392,34,34,3253,174
rovers,40,143,53,53,933,71
satellite,36,158,56,56,3314,54
scanalyzer-3d,20,118,37,62,14798,23
storage,20,54,22,22,596,27
tetris,19,25,61,119,27598,2009
thoughtful,15,47,75,75,4342,234
tidybot,17,58,56,56,38100,427
tpp,30,58,92,92,1798,99
transport,20,87,124,2031,14590,20
trucks,18,69,35,35,2695,20
woodworking,30,130,47,898,2899,172
zenotravel,19,53,26,26,1738,14
Total,950,3826,51,81,2053,50
\end{filecontents*}
\pgfplotstabletypeset[
    column name={},
    column type=l,
    every head row/.style={
    before row={
    \toprule
    \multirow{2}{*}{Domains} & \multirow{2}{*}{Problems} & \multirow{2}{*}{Plans} & \multirow{2}{*}{Plan Size} & \multirow{2}{*}{Plan Cost} &  \multirow{2}{*}{Operators} & \multirow{2}{*}{Variables} \\
    },
    after row=\midrule,
    },
    every last row/.style={before row=\toprule,after row=\midrule},
    columns/domain/.style ={column type/.add={@{\hspace{2pt}}}{}},
    columns/problems/.style ={column type=r, column type/.add={@{\hspace{0pt}}}{}},
    columns/plans/.style={column type=r, column type/.add={@{\hspace{9pt}}}{}},
    columns/operators/.style={column type=r, column type/.add={@{\hspace{9pt}}}{}},
    columns/initialsize/.style={column type=r, column type/.add={@{\hspace{9pt}}}{}},
    columns/initialcost/.style={column type=r, column type/.add={@{\hspace{9pt}}}{}},
    columns/variables/.style={column type=r, column type/.add={@{\hspace{9pt}}}{}},
    col sep=comma,
    string type,
    ignore chars={"}
     ]{media/domain_info1.csv}
\end{table}

We calculate the flexibility of a plan, referred to as $\mathit{flex}$ \cite{siddiqui_patrik_2012}, by computing the ratio of operator pairs with no basic or transitive ordering to the total number of operator pairs. Given a POP $\pi_{pop}=\langle \mathcal{O}, \prec\rangle$, 
    \begin{equation}
    \mathit{flex}(\pi_{pop}) = 1- \frac{|\prec|}{\Sigma_{i=1}^{|\mathcal{O}|-1}i}
    \end{equation} 

The denominator $\Sigma_{i=1}^{|\mathcal{O}|-1}i$, equivalent to $\binom{|\mathcal{O}|}{2}$, calculates the total number of pairs that can be formed from a set of  $|\mathcal{O}|$ elements. \emph{flex} is a measure of how many ordering constraints there are in the POP, normalized by the total number of potential ordering constraints \cite{maxsat}, and its value ranges from 0 to 1. The higher the flex, the more flexible the POP is.

We also extend our experimental study by applying MaxSAT reordering techniques, particularly MaxSAT encoding by \citeauthor{maxsat} (\citeyear{maxsat}), and by \citeauthor{maxsat_reinst} (\citeyear{maxsat_reinst}), on the dataset presented in Table \ref{tbl:dataset}. To solve their encodings, we have used \emph{Loandra} \cite{loandra} with MaxPre \cite{maxpre} as a preprocessing tool. Loandra is an anytime MaxSAT solver that incrementally improves solution quality over time. It is built on top of OpenWBO \cite{Open-WBO}, leveraging its core MaxSAT solving capabilities while adding anytime behavior. Notably, to improve the performance of MaxSAT techniques, we also include the operator symmetry-breaking constraints \cite{say_optimize_pop}. 

We use notations for different methodologies to describe our experimental findings clearly. We denote block deordering as BD, and specify MaxSAT compilation by \citeauthor{maxsat} (\citeyear{maxsat}), and by \citeauthor{maxsat_reinst} (\citeyear{maxsat_reinst}) as MR, and MRR, respectively.  \emph{FIBS} refers to the basic compilation of our algorithm, which does not eliminate redundant operators. We use the Wilcoxon signed-rank test \cite{Wilcoxon1992} for comparative analysis. 
This is a non-parametric statistical test used to determine whether there is a significant difference between paired observations of two methods. The test statistic produced by this test is converted into a p-value. A p-value below 0.05 indicates a significant difference between the outcomes of the two methods. We use asterisks(*) for the significance levels of p-values: one asterisk(*) for 0.05 to >0.01, and two asterisks(**) for < 0.001. All experiments were conducted on an 8-core, 2.80GHz Core i7-1165G7 CPU, with a 30-minute time limit. 

The source code and datasets used for this experiment are publicly available at \href{https://github.com/sabah0312/fast-downward-pd}{github.com/sabah0312/fast-downward-pd} and \href{https://github.com/sabah0312/IPC-Solutions}{github.com/sabah0312/IPC-Solutions}. This ensures full reproducibility of all reported results.

\subsection{Effectiveness of the FIBS Algorithm}
FIBS integrates EOG and BD with block-substitution, improving plan flexibility in a substantial number of plans over EOG or BD individually. We estimate plan flexibility after performing each of the four phases in FIBS: 1) EOG, 2) $1^{st}$ Subtitution-Deorder (SD1), 3) BD, and 4) $2^{nd}$ Subtitution-Deorder (SD2). 

Table \ref{tbl:table1}  summarizes the experimental results of FIBS. For each phase, we present the arithmetic mean of plan flex ($flex$), CPU execution time ($T$ in seconds), and the number of plans (\#) that exhibited improved flex compared to the preceding phase. Compared to EOG, FIBS exhibits greater flex across 30 domains, leading to a 62\% overall improvement in flex. The SD1, BD, and SD2 phases improved $\mathit{flex}$ in a substantial number of plans, specifically 427, 2324, and 425 plans, respectively. It is worth noting that in this experiment, we ensured that plan flexibility was not compromised in any phase.

\begin{table}[!tbp]
\caption{{Experimental results of FIBS algorithm. $flex$ and $T$ present the arithmetic mean flex and the CPU execution time in seconds, respectively, for the corresponding phases (EOG, SD1, BD, and SD2) in FIBS. The column \# shows the number of plans with improved flex for a phase compared to its preceding one. Asterisks indicate significance levels from paired Wilcoxon signed-rank tests comparing $\mathit{flex}$ values of a phase to its preceding phase. Empty cells indicate no difference.}}
    \label{tbl:table1}
    \centering
    \begin{filecontents*}{media/nbsd_result.csv}
"domain","feog","teog","fbsd1","nbsd1","tbsd1","fbd","nbd","tbd","fbsd2","nbsd2","tbsd2"
"barman","0.004","7","","","1200","0.167","32","68","0.17","*12","88"
"blocks","0","23","","","541","0.14","172","53","0.143","**37","16"
"child-snack","0.695","<1","","","1","0.842","8","2","0.843","1","1"
"depots","0.265","<1","0.266","4","30","0.333","62","11","","","2"
"elevator","0.06","<1","","2","32","0.23","605","8","0.234","**226","9"
"floor-tile","0.147","<1","","3","1","0.196","41","<1","","1","<1"
"freecell","0.07","<1","","","25","0.078","111","2","","*8","11"
"genome-edit-distances","0.005","3","","","332","0.32","87","33","0.324","**23","2"
"grid","0","<1","","","14","0.017","4","1","","","8"
"gripper","0.017","<1","","","48","0.713","19","4","","","<1"
"hiking","0.056","<1","0.063","**69","2","0.074","62","<1","0.075","*7","1"
"logistics","0.576","1","","**17","18","0.591","99","14","0.592","**18","7"
"mystery","0.123","<1","","","<1","","","<1","","","<1"
"mystery-prime","0.176","<1","0.181","5","<1","0.182","1","<1","","","<1"
"no-mystery","0.037","<1","","","<1","0.119","26","<1","","","<1"
"parc-printer","0.629","<1","","","1","","","<1","","","1"
"parking","0.003","1","","","36","0.046","79","1","","1","23"
"pathways","0.493","2","0.497","**20","197","0.551","32","234","0.552","2","33"
"peg-solitaire","0","<1","","","<1","0.232","243","<1","0.233","**29","14"
"pipesworld","0.171","<1","0.178","**95","11","0.193","95","1","","*11","4"
"rovers","0.718","1","","**17","16","0.775","132","82","","*13","8"
"satellite","0.438","4","0.441","**37","195","0.513","81","12","","*6","14"
"scanalyzer-3d","0.068","1","","","2","0.561","105","<1","0.565","*11","<1"
"storage","0.12","<1","","","1","0.373","46","<1","","","<1"
"tetris","0.536","1","","","2","0.549","13","1","0.551","2","2"
"thoughtful","0.085","1","","","133","0.086","6","1","","","132"
"tidybot","0.067","2","","","47","0.082","15","<1","","","8"
"tpp","0.315","1","0.319","**31","371","0.477","43","21","0.478","5","11"
"transport","0.462","8","","2","292","0.474","51","101","0.475","*8","34"
"trucks","0.025","<1","","","<1","0.038","29","<1","","","<1"
"woodworking","0.907","<1","0.932","**122","4","","","<1","","","<1"
"zenotravel","0.389","<1","","3","<1","0.407","25","<1","0.41","4","<1"
"Total","0.2","2.33","0.213","**427","95","0.323","2324","19","0.325","**425","106"
\end{filecontents*}
\pgfplotstabletypeset[
    column type=l,
    every head row/.style={
    before row={
    \toprule
    \multirow{2}{*}{Domains}& \multicolumn{2}{c}{EOG} &    \multicolumn{3}{c}{SD1} & \multicolumn{3}{c}{BD} & \multicolumn{3}{c}{SD2}\\
    \cmidrule(lr){2-3}\cmidrule(lr){4-6}\cmidrule(lr){7-9}\cmidrule(lr){10-12}
    },
    after row=\midrule,
    },
    every last row/.style={before row=\toprule,after row=\midrule},
    display columns/0/.style={column type={p{.22\textwidth}}},
    columns/domain/.style ={column name=},
    columns/nbsd1/.style ={column name=$\#$, column type=r},
    columns/tbsd1/.style ={column name=$T$, column type=r},
    columns/nbd/.style={column name=$\#$, column type=r},
    columns/tbd/.style={column name=$T$, column type=r},
    columns/nbsd2/.style ={column name=$\#$, column type=r},
    columns/tbsd2/.style ={column name=$T$, column type=r},
    columns/teog/.style={column name=$T$, column type=r},
    columns/feog/.style={column name=$flex$, column type=r},
    columns/fbsd1/.style={column name=$flex$, column type=r, column type/.add={@{\hspace{18pt}}}{}},
    columns/fbd/.style={column name=$flex$, column type=r, column type/.add={@{\hspace{18pt}}}{}},
    columns/fbsd2/.style={column name=$flex$, column type=r, column type/.add={@{\hspace{18pt}}}{}},
    col sep=comma,
    string type,
    ignore chars={"},
    ]{media/nbsd_result.csv}
\end{table}

The second phase, SD1, successfully eliminates orderings in 11\% of plans across 14 domains. This improvement is particularly notable in hiking, mystery-prime, pathways, pipesworld,  satellite, tpp, 
 and woodworking domains. Then, block deordering enhances plan flexibility in 60\% of the plans over SD1 across 29 domains.  
 The subsequent phase, SD2, also eliminates orderings in 11\% of the total plans after performing block deordering across 20 domains, notably in barman, blocks, elevator, freecell, genome-edit-distances, hiking, logistics, peg-solitaire, pipesworld, rovers, satellite, scanalyzer-3d, and transport.

 The results show a general upward trend in $\mathit{flex}$ across phases. Though each phase improves $\mathit{flex}$ in a substantial number of plans, the actual improvement in $\mathit{flex}$ is comparatively less in SD1 and SD2 than in phase BD. On average, $\mathit{flex}$ increases from 0.20 in the initial EOG phase to 0.213 in SD1, 0.323 in BD, and 0.325 in SD2. One reason is that block-substitution can improve flexibility when additional resources (e.g., elevators, robots, cars, etc.) are available in the corresponding problem task. Besides,  SD1 only attempts to replace primitive blocks (i.e., operator) to improve flexibility, and then SD2 considers only blocks for replacement instead of blindly searching for suitable subplans. Therefore, other subplans, not enclosed by blocks within a BDPO plan, remain unexplored in the FIBS algorithm. Our experimental results demonstrate that block substitution can enhance $\mathit{flex}$ in plans, and FIBS performance could be further improved by exploring more subplans for substitution.

   \begin{figure}[!tbp]
     \centering
     \includegraphics[width=\linewidth]{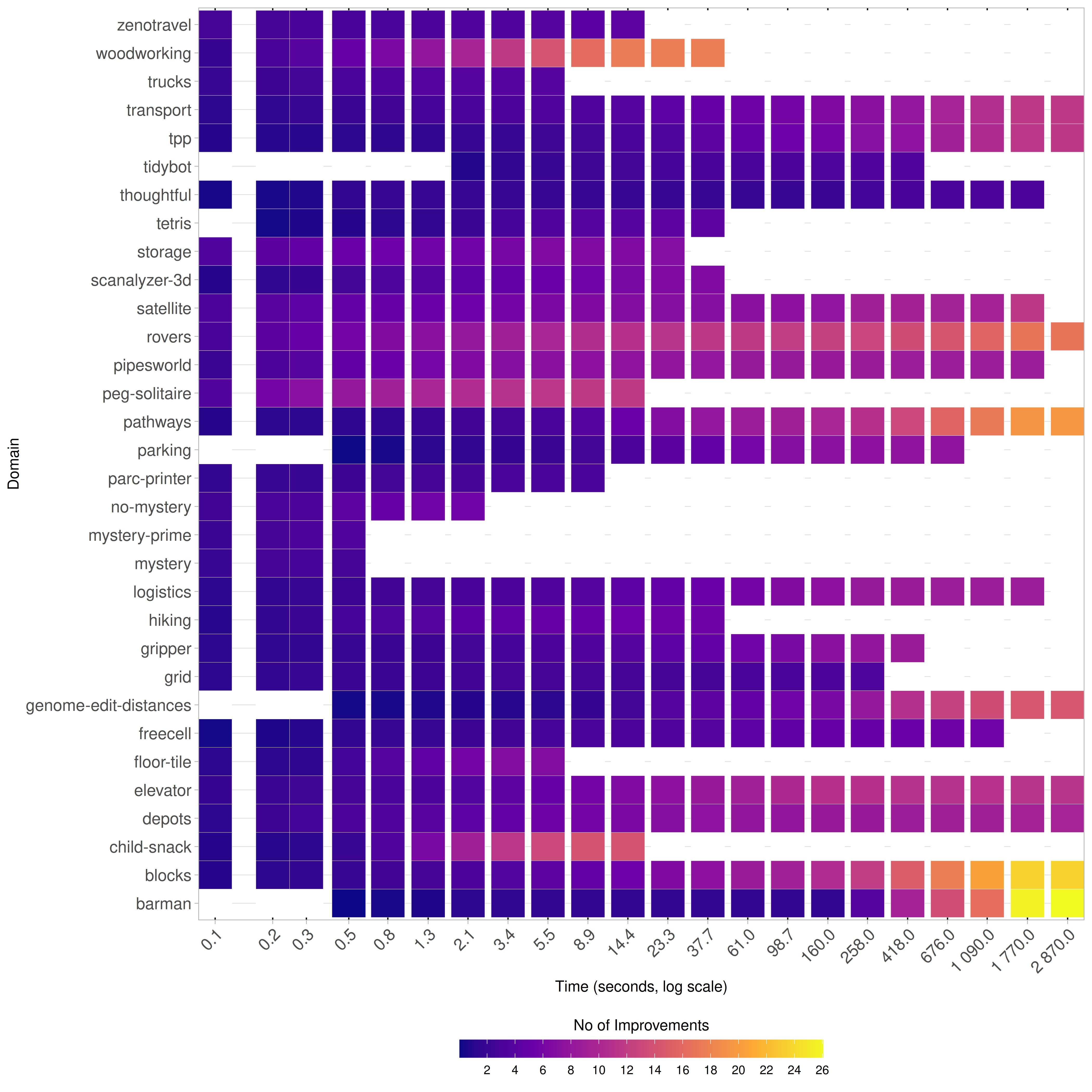}
     \caption{Heatmap of the average number of improvements in flexibility per plan, achieved by FIBS, across domains over time.  The color intensity encodes the arithmetic average number of flex improvements achieved within the given time for each domain. Darker purple indicates fewer improvements, while yellow corresponds to higher improvements.}
     \label{fig:anytime_heatmap}
 \end{figure}

We also illustrate how the FIBS algorithm improves the flexibility of plans across different domains over time in the heatmap, presented in Figure~\ref{fig:anytime_heatmap}. In this plot, the x-axis represents time in seconds, and the y-axis lists the planning domains. The color intensity indicates the arithmetic average number of flex improvements achieved within the given time for each domain. Here, the number of flex improvements corresponds to the number of successful ordering removals achieved by the SD1, BD, or SD2 phases in FIBS. Darker purple indicates fewer improvements, while yellow corresponds to higher improvements.  Overall, this figure illustrates the anytime nature of FIBS, demonstrating its ability to provide incremental improvements as computation time increases.

The visualization in Figure \ref{fig:anytime_heatmap} also highlights the domain-specific impact of FIBS, showcasing both its effectiveness and its limitations. In most domains, FIBS exhibits early improvements through block deordering (BD), which are then incrementally extended over time through block substitution (i.e., the SD2 phase). Domains such as barman and blocks exhibit sustained increases in the average number of improvements, even at later time periods, demonstrating FIBS’s capacity to improve flexibility when additional time is allowed continuously. Other domains, including genome-edit-distance, pathways, pipesworld, rovers, and satellite, exhibit moderate improvements that plateau earlier. Additionally, certain domains, such as barman, parking, and tidybot, experience late improvements, primarily due to a lack of enhancements in the primary block substitution phase (SD2 phase). Notably, some domains, such as child-snack, floor-tile, hiking, peg-solitaire, logistics, trucks, and woodworking, show only minimal benefits, with improvements stagnating very early. This indicates that FIBS’s impact is domain-dependent, often constrained by the structural characteristics of the planning tasks. Another contributing factor is that FIBS’s performance can be limited by the number of blocks provided by block deordering, which reduces its scope for subsequent improvements, as discussed earlier. 
 
\subsection{Comparative Analysis of Flexibility Improvement via Block-Substitution (FIBS) and MaxSAT-based Reordering}

We conduct an extensive comparative study of the MaxSAT reordering techniques (MR and MRR) and the FIBS algorithm. We use the solutions generated by EOG as a synthetic baseline due to its trivial execution time. When MR or MRR fail to produce a solution for a given plan, we provide them with the solution generated by EOG to ensure a fair comparison.

\begin{table}[!tbp]
    \caption{{The experimental results of  MR, MRR, and FIBS. $flex$ and $T$ present the arithmetic mean flex, and mean execution time in seconds, respectively,
for the respective algorithm.}}
    \label{tbl:table3}
    \centering
    \begin{filecontents*}{media/bsd_mr_mrr2.csv}
"domain","mrflex","mrtime","mrrflex","mrrtime","bsdflex","bsdtime"
"barman",0,1517.81,"0",1696.04,"**0.17",1383.65
"blocks",0,921.63,"0",1483.53,"**0.17",245.51
"child-snack",0.73,363.57,"0.85",1678.02,"0.84",3.86
"depots",0.31,288.03,"0.38",1649.64,"0.34",19.3
"elevator",0.15,131.27,"0.16",612.94,"**0.25",81.51
"floor-tile",0.16,1.96,"0.15",485.04,"**0.2",1.67
"freecell",0.08,20.47,"**0.12",1392.28,"0.08",27.16
"genome-edit-distances",0.01,1604.22,"0.01",1795.25,"**0.32",374.92
"grid",0,7.06,"0",929.01,"0.02",13.08
"gripper",0.02,1255.13,"0.02",1501.22,"**0.71",58.26
"hiking",0.06,19.76,"0.11",1591.42,"0.07",3.03
"logistics",0.59,95.98,"0.61",1185.8,"0.59",42.73
"mystery",0.1,216.47,"*0.15",87.69,"0.1",0.06
"mystery-prime",0.18,59.56,"**0.24",375.29,"0.18",0.1
"no-mystery",0.04,83.41,"0.04",951.99,"**0.12",0.52
"parc-printer",0.63,1.07,"0.63",1185.72,"0.63",1.17
"parking",0,21.07,"0",1726.02,"**0.05",62.18
"pathways",0.51,432.08,"0.51",334.96,"**0.53",417.62
"peg-solitaire",0,230.22,"0",822.7,"**0.23",14.71
"pipesworld",0.18,41.4,"**0.24",959.39,"0.19",15.17
"rovers",0.71,205.87,"0.75",557.66,"**0.76",76.94
"satellite",0.44,263.36,"0.5",1406.42,"0.51",225.39
"scanalyzer-3d",0.08,541.68,"0.13",1497.02,"**0.57",3.88
"storage",0.12,35.55,"0.12",2.29,"**0.37",0.88
"tetris",0.54,3.67,"0.56",1732.84,"0.55",5.98
"thoughtful",0.09,3.62,"*0.1",1085.76,"0.09",266.78
"tidybot",0.07,4.59,"0.07",35.63,"**0.08",58.7
"tpp",0.4,631.67,"0.4",668.91,"**0.46",434.1
"transport",0.47,350.89,"*0.5",1009.44,"0.47",438.47
"trucks",0.03,81.43,"0.03",0.88,"**0.04",0.57
"woodworking",0.91,1.27,"0.93",1340.6,"0.93",4.1
"zenotravel",0.39,2.32,"0.39",2.29,"**0.41",1
"Total",0.23,241.88,"0.25",1038.66,"**0.32",105.82
\end{filecontents*}
\pgfplotstabletypeset[
    column type=l,
    every head row/.style={
    before row={
    \toprule
    \multirow{2}{*}{Domains}& \multicolumn{2}{c}{MR} &    \multicolumn{2}{c}{MRR} & \multicolumn{2}{c}{FIBS} \\
    \cmidrule(lr){2-3}\cmidrule(lr){4-5}\cmidrule(lr){6-7}
    },
    after row=\midrule,
    },
    every last row/.style={before row=\toprule,after row=\midrule},
    display columns/0/.style={column type={p{.24\textwidth}}},
    columns/domain/.style ={column name=},
    columns/bsdflex/.style ={column name=$flex$, column type=r, column type/.add={@{\hspace{20pt}}}{}},
    columns/bsdtime/.style ={column name=$T$, column type=r},
    columns/mrflex/.style={column name=$flex$, column type=r},
    columns/mrtime/.style={column name=$T$, column type=r},
    columns/mrrflex/.style ={column name=$flex$, column type=r, column type/.add={@{\hspace{20pt}}}{}},
    columns/mrrtime/.style ={column name=$T$, column type=r},
    col sep=comma,
    string type,
    ignore chars={"},
    ]{media/bsd_mr_mrr2.csv}
\end{table}

Table \ref{tbl:table3} succinctly presents the experimental results of MR, MRR, and FIBS.  FIBS outperforms MR in 25 domains. Compared to MRR, FIBS provides better flex in 19 domains. Among these domains, FIBS produces better solutions in a significant number of plans in 17 domains, such as barman, blocks, elevator, floor-tile, genome-edit-distances, gripper, no-mystery, and parking. On the other hand, MRR surpasses FIBS in  11 domains, such as child-snack, freecell, logistics, mystery, mystery-prime, and pipesworld.  Overall, FIBS exhibits the highest average flex of 0.32, followed by MRR and MR with mean flexes of 0.25 and 0.23, respectively. 

The main drawbacks of MaxSAT reorderings are their high computation time. MR and MRR encodings require significantly prolonged computational time, and often do not produce any solution for a plan. In our experiment, MR achieves an 86\% coverage with a mean execution time of 242 seconds. MR becomes infeasible for large plans (typically more than 200 operators) because of the size of transitivity clauses \cite{maxsat}. The encoding size escalates in MRR due to additional formulas, reducing its coverage from 85\% to 54\%, and increasing the mean run time to 1039 seconds. In contrast, FIBS, which iteratively and locally enhances flexibility,  takes an average execution time of 106 seconds with full coverage.

 \begin{figure}[!tb]
     \centering
     \includegraphics[width=.8\linewidth]{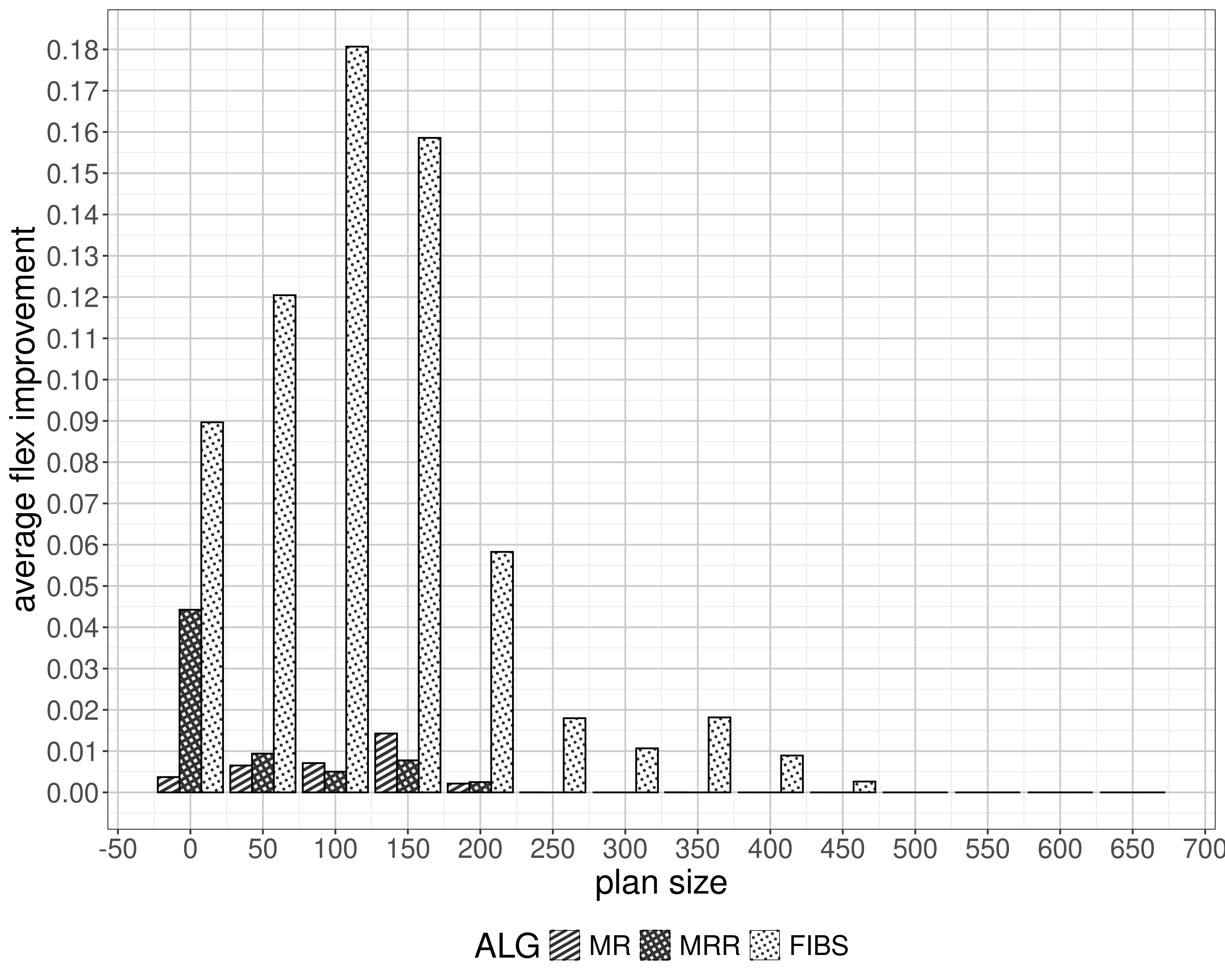}
     \caption{Arithmetic average of flex improvement of MR, MRR, and FIBS compared to EOG as a function of plan size}
     \label{fig:size-flex}
 \end{figure}
 
 Figure \ref{fig:size-flex} illustrates the arithmetic average of flex improvement of MR, MRR, and FIBS compared to EOG. The x-axis represents the plan size, segmented into intervals (e.g., 0–50, 51–100, etc.), while the y-axis denotes the corresponding average flexibility improvement compared to EOG. Distinct bar patterns distinguish the algorithms: MR (striped), MRR (dense cross-hatch), and FIBS (light cross-hatch). Each bar group shows the comparative performance of the algorithms within a specific plan size range. The FIBS algorithm consistently demonstrates superior performance over MR and MRR, especially for small to medium-sized plans. Notably, for plan sizes between 100 and 150, FIBS achieves the highest average flexibility improvement, peaking around 0.18. In contrast, MR and MRR  yield considerably lower improvements, rarely exceeding 0.02. The effectiveness of all algorithms declines with increasing plan size, with improvements becoming negligible beyond a plan size of 250. This trend indicates that while FIBS significantly enhances execution flexibility for smaller plans, its impact diminishes for larger and more complex plans, possibly due to the overhead introduced by its substitution operations or limitations in identifying suitable candidate subplans for substitutions as plan size grows. Overall, this plot illustrates the scalability limitations of the evaluated algorithms and demonstrates that FIBS enhances plan flexibility in small to moderately sized planning problems.

  \begin{figure}[!tb]
     \centering
     \includegraphics[width=.8\linewidth]{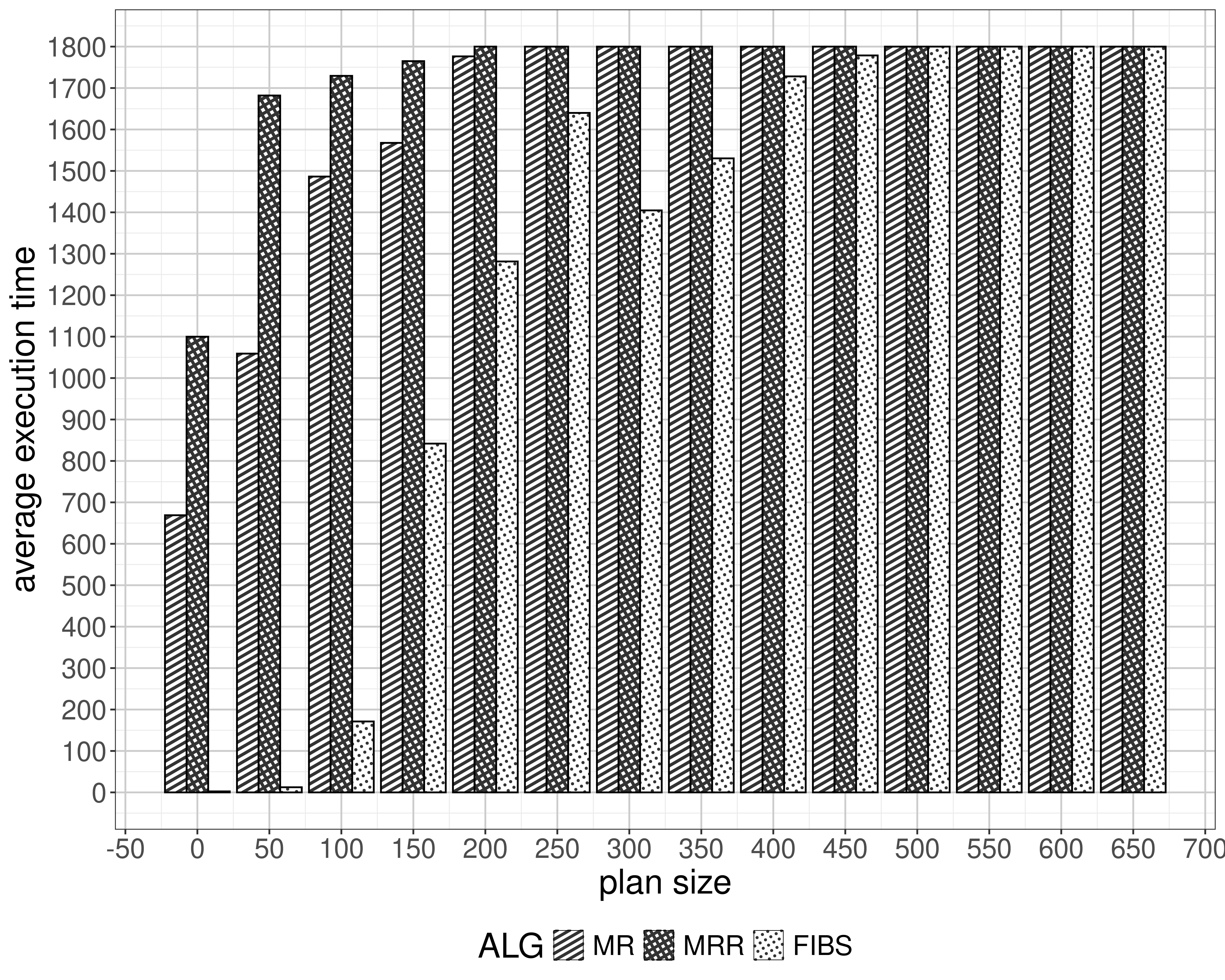}
     \caption{Arithmetic average of execution time flex of MR, MRR, and FIBS as a function of plan size}
     \label{fig:time-flex}
 \end{figure}

Similarly, Figure \ref{fig:time-flex} presents the arithmetic average of execution time flexibility for the MR, MRR, and FIBS algorithms as a function of plan size. Here, the y-axis measures the average execution time, ranging from 0 to 1800 seconds, where 1800 seconds corresponds to the imposed time limit. The execution time of all three algorithms increases with plan size, but at different rates.  FIBS exhibits a steady, linear increase, while MR and MRR experience much steeper rises, rapidly reaching the 1800-second ceiling. Specifically, MRR saturates first, between plan sizes 200 and 250, followed by MR between 250 and 300. By contrast, FIBS consistently achieves lower execution times than both MR and MRR across all plan sizes and only approaches the time limit much later, between plan sizes 500 and 550. After this point, all three algorithms plateau at the upper bound set by the cutoff. This plot highlights the scalability and robustness of FIBS compared to MR and MRR.

 We also employ FIBS on the solutions of MR and MRR to evaluate the enhancement in plan flexibility resulting from the integration of FIBS with MaxSAT reorderings. We refer to the methodologies where FIBS is applied to the solutions of EOG, MR, and MRR as EOG+FIBS, MR+FIBS, and MRR+FIBS, respectively.

\subsection{Performance Analysis of FIBS on Different Base Deordering/Reordering Strategies}
We also employ FIBS on the POPs generated by MR and MRR encodings instead of EOG to investigate the potential increase in plan flexibility achieved by combining FIBS with MaxSAT reorderings. We denote the compilation as \emph{EOG + FIBS} when FIBS is employed on the deordered plans generated by EOG. Similarly, \emph{MR + FIBS} and \emph{MRR + FIBS} refer to the compilations where FIBS is applied on the POPs generated by MR and MRR, respectively. 

 \begin{table}[!tbp]
 \caption{Comparison of FIBS on different base methods(EOG, MR and MRR). For each method, $S$ and $=$ indicate the number of solutions provided, and the number of solutions that have the highest flexibility, while $>$ indicates the number of solutions that have strictly better flexibility than any other method.}\label{tbl:fibs2}
    \centering
    \begin{filecontents*}{media/bsd_mr_mrr.csv}
"domain","bsd1","bsd2","bsd3","bsdmr1","bsdmr2","bsdmr3","bsdmrr1","bsdmrr2","bsdmrr3"
"barman",54,36,"36","11","4","4","","",""
"blocks",234,90,"79","99","15","6","59","11","3"
"child-snack",8,3,"3","8","","","8","3","3"
"depots",73,2,"1","70","9","8","33","12","11"
"elevator",742,187,"186","101","18","17","35","8","8"
"floor-tile",44,6,"6","13","1","1","1","1","1"
"freecell",208,17,"9","189","28","20","80","39","38"
"genome-edit-distances",93,20,"11","81","9","","5","",""
"grid",15,1,"","14","2","1","12","4","3"
"gripper",20,20,"1","19","19","","5","5",""
"hiking",111,12,"12","111","","","37","8","8"
"logistics",161,32,"10","150","58","36","44","27","16"
"mystery",28,12,"","25","11","","28","19","7"
"mystery-prime",61,22,"","61","22","","57","33","13"
"no-mystery",28,6,"2","28","7","3","24","4","2"
"parc-printer",116,54,"29","116","81","1","113","79",""
"parking",144,35,"21","144","19","5","","",""
"pathways",36,22,"22","21","","","28","",""
"peg-solitaire",258,25,"6","258","34","13","158","27","17"
"pipesworld",392,67,"2","388","71","3","266","100","55"
"rovers",143,24,"15","125","26","2","132","43","19"
"satellite",158,34,"25","146","14","5","73","21","17"
"scanalyzer-3d",118,19,"16","118","11","8","35","10","7"
"storage",54,20,"20","","","","","",""
"tetris",25,11,"11","25","7","7","4","1","1"
"thoughtful",47,6,"6","47","4","4","17","5","5"
"tidybot",58,17,"17","","","","","",""
"tpp",58,18,"12","41","17","3","42","15","1"
"transport",87,41,"35","50","12","6","18","8","5"
"trucks",69,14,"2","67","16","","69","16",""
"woodworking",130,21,"15","130","4","","100","57","51"
"zenotravel",53,19,"19","","","","","",""
"Total",3826,913,"629","2656","519","153","1483","556","291"
\end{filecontents*}
\pgfplotstabletypeset[
    column type=l,
    every head row/.style={
    before row={
    \toprule
    \multirow{2}{*}{Domains}  & \multicolumn{3}{c}{EOG+FIBS} & \multicolumn{3}{c}{MR+FIBS}& \multicolumn{3}{c}{MRR + FIBS} \\
    \cmidrule(lr){2-4}\cmidrule(lr){5-7}\cmidrule(lr){8-10}
    },
    after row=\midrule,
    },
    every last row/.style={before row=\toprule,after row=\midrule},
    columns/domain/.style ={column name=, column type=l},
    columns/bsd1/.style ={column name=$S$, column type=r},
    columns/bsd2/.style ={column name={=}, column type=r},
    columns/bsd3/.style={column name=$>$, column type=r},
    columns/bsdmr1/.style ={column name=$S$, column type=r,  column type/.add={@{\hspace{20pt}}}{}},
    columns/bsdmr2/.style ={column name={=}, column type=r},
    columns/bsdmr3/.style={column name=$>$, column type=r},
    columns/bsdmrr1/.style ={column name=$S$, column type=r, column type/.add={@{\hspace{20pt}}}{}},
    columns/bsdmrr2/.style ={column name={=}, column type=r},
    columns/bsdmrr3/.style={column name=$>$, column type=r},
    col sep=comma,
    string type,
     ignore chars={"},
    ]{media/bsd_mr_mrr.csv}
\end{table}
Table \ref{tbl:fibs2} presents a comparative analysis of FIBS when applied on top of the different base deordering/reordering methods: EOG, MR, and MRR. For each configuration, the table shows the number of solutions generated ($S$), the number of solutions that achieve the highest flexibility ($=$), and those that are strictly better than any other method ($>$). Across all domains, FIBS combined with EOG generates the highest number of solutions with highest flex, followed by MRR and MR. In terms of strictly superior flexibility, EOG+FIBS also outperforms the others, contributing 629 solutions that are better than any other method, compared to 153 for MR+FIBS and 291 for MRR+FIBS. Specifically, EOG+FIBS provides more plans with strictly better $\mathit{flex}$ in domains such as barman, blocks. On the other hand, MRR+FIBS finds more strictly better plans in depots, freecell, mystery-prime, peg-solitaire, pipesworld, rovers, and woodworking domains, and MR+FIBS achieves the best result on this measure in logistics domains.

\begin{figure}[!tb]
    \begin{subfigure}[t]{0.48\linewidth}
        \includegraphics[width=\linewidth]{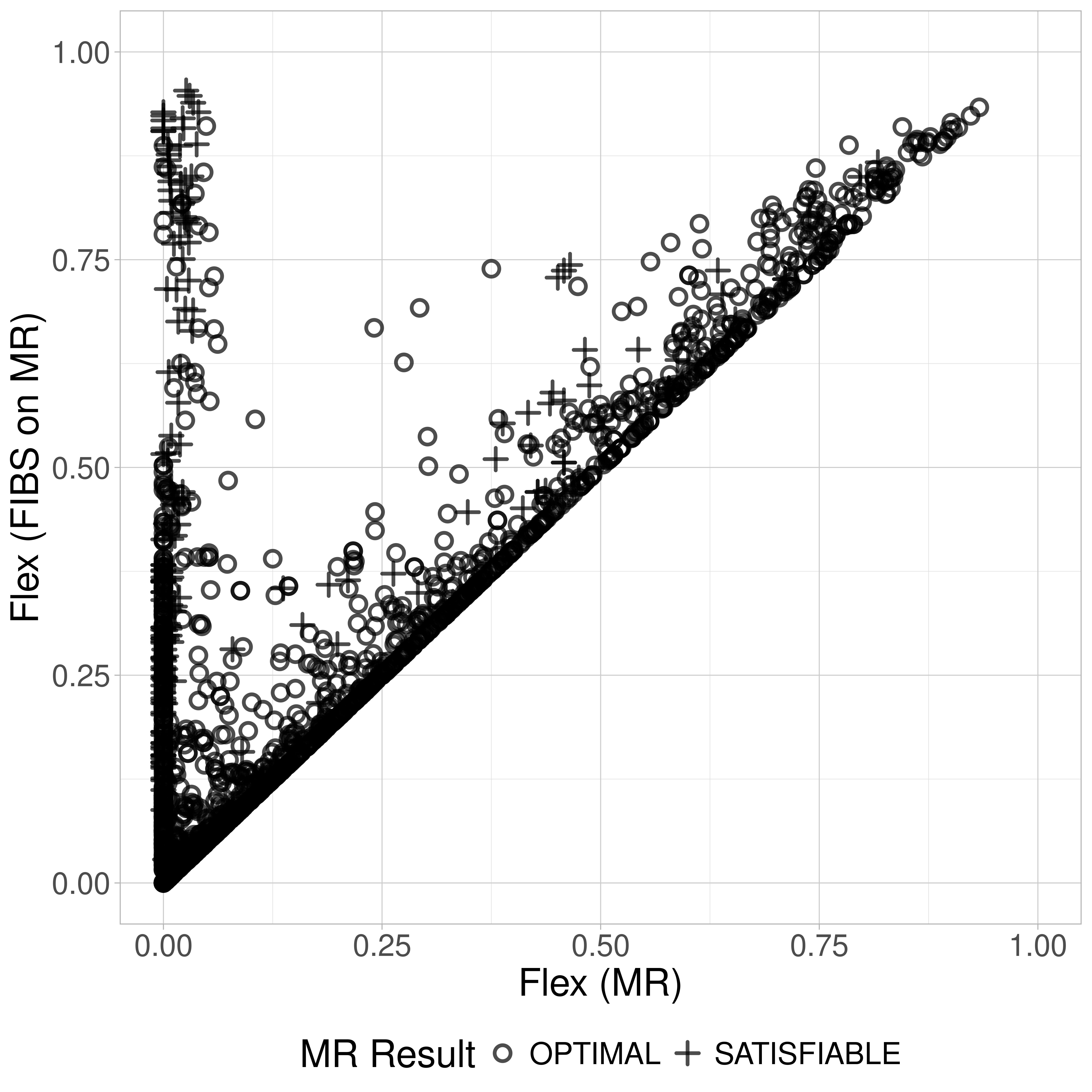}
        \caption{}
    \end{subfigure}
    \hfill
    \begin{subfigure}[t]{0.48\linewidth}
        \includegraphics[width=\linewidth]{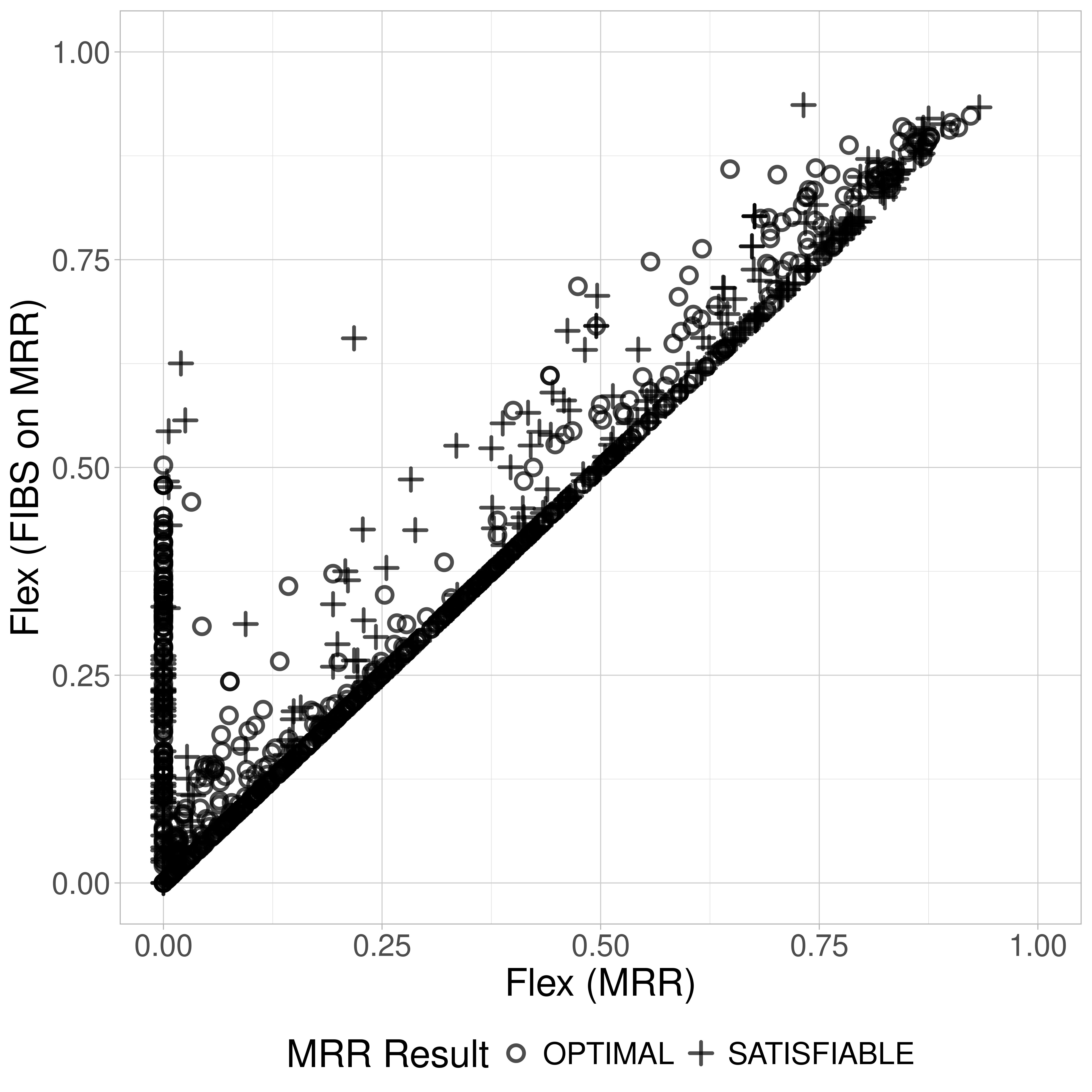}
        \caption{}
    \end{subfigure}%
    \caption{Comparing (a) $\mathit{flex}$ of employing FIBS on MR solutions to that of MR solutions. (b) $\mathit{flex}$ of employing FIBS on MRR solutions to that of MRR solutions.}
    \label{fig:scatter_mr}
\end{figure}

In Figure \ref{fig:scatter_mr}, we illustrate the flexibility improvements that FIBS provides over the optimal and satisfiable solutions of MR and MRR. In each plot, the x-axis reports the flexibility of the solutions of  MaxSAT reoderings (MR or MRR), while the y-axis shows the flexibility obtained when FIBS is applied on the same reordered plan. Circles denote optimal solutions, and plus signs represent satisfiable solutions.
These plots clearly demonstrate that the integration of FIBS effectively enhances the plan flexibility for both optimal and satisfiable solutions of MaxSAT reorderings.

 \begin{figure}
     \centering
     \includegraphics[width=.9\linewidth]{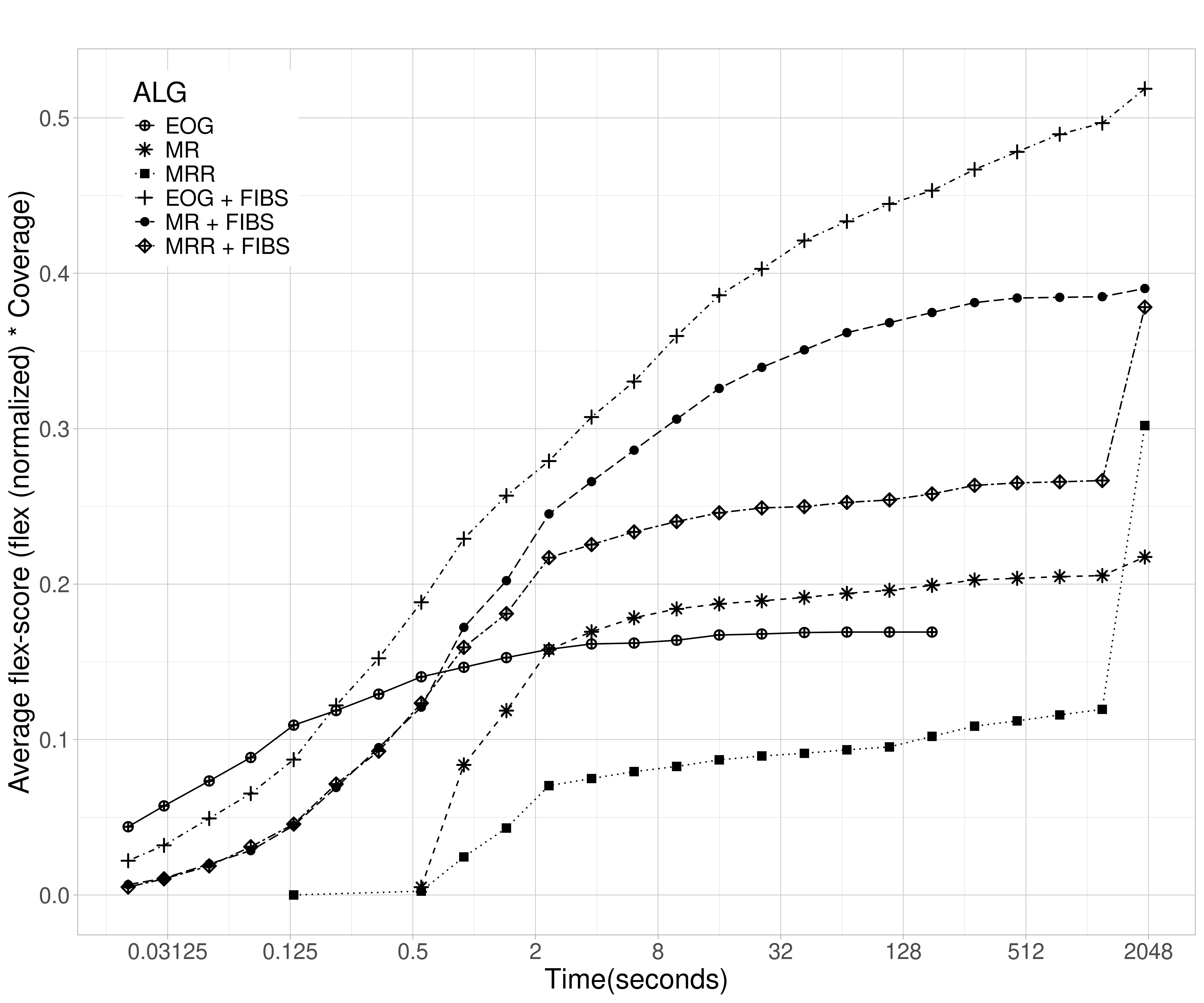}
     \caption{Average $\mathit{flex}{-}score$ as a function of time after performing EOG, MR, MRR, EOG + FIBS (FIBS on EOG),  MR+FIBS (FIBS on MR), and MRR+ FIBS (FIBS on MRR) for plans from IPC domains. $\mathit{flex}$ is normalized to the interval between the lowest and highest $\mathit{flex}$ found for a problem. The x-axis is displayed in the log scale. The spikes in MR, MRR, MR +FIBS and MRR + FIBS performance curves are due to their satisfiable solutions provided by the end of the time limit.}
     \label{fig:time-flex2}
 \end{figure}

Figure \ref{fig:time-flex2} depicts a comparison among EOG, MR, MRR, EOG+FIBS, MR+FIBS and MRR+FIBS  based on their $\mathit{flex}{-}score$ over time. The $\mathit{flex}{-}score$ is estimated by multiplying the average normalized flex by the coverage within a specified time frame. Here, coverage represents the proportion of plans for which the respective method has completed computation out of the total number of plans. The plot shows that EOG starts providing solutions in under 0.03 seconds, whereas MR and MRR take nearly 0.5 seconds to begin. EOG's performance plateaus after approximately 8 seconds, reaching an average  $\mathit{flex}{-}score$ close to 0.15. EOG provides solutions for all plans within 128 seconds. MR, EOG+FIBS, and MR+FIBS intersect EOG's performance line around 0.5 seconds, while MRR+FIBS intersects it after 2 seconds. EOG+FIBS consistently improves plan flexibility over time, reaching the highest  $\mathit{flex}{-}score$ of 0.5. 

It's important to note that MR and MRR yield satisfiable solutions at the end of the time limit, resulting in spikes in their performance lines near the end. While 87\% of MR solutions are optimal, MRR provides optimality in only 38\% of its solutions. This difference explains why the spike in the lines of MRR and MRR+FIBS near the end of the time limit is more drastic than that of MR and MR+FIBS.

\subsection*{Results of Eliminating Redundant Operators}
\begin{table}[!tbp]
\caption{Experimental results of eliminating redundant operators in FIBS using Backward and Greedy Justification, compared with Minimal Plan Reduction (MPR). FIBS with backward justification is denoted as $FIBS\_BJ$, and FIBS with greedy justification as $FIBS\_GJ$. Domain-wise, the column \emph{$c_{initial}$} shows the average initial plan cost, and $\Delta c$ specifies the geometric mean reduction in percentage in plan cost for each method compared to the initial cost. Asterisks indicate statistical significance levels from a paired Wilcoxon signed-rank test on plan cost between FIBS and MPR.}
    \label{tbl:table2}
    \centering
    \begin{filecontents*}{media/bsd_rr.csv}
"domain","ceog","bj","fj","kr"
"barman",288,"2.5","5.2","*7.55"
"blocks",60,"9.23","14.65","14.92"
"child-snack",67,"6.8","9.77","9.27"
"depots",46,"6.07","7.03","**10.31"
"elevator",69,"5.54","**5.76","4.36"
"floor-tile",81,"3.25","5.09","5.52"
"freecell",50,"5.35","5.39","5.29"
"genome-edit-distances",33,"2.09","**2.09",""
"grid",53,"4.81","4.81","4.81"
"gripper",59,"4.32","4.32","4.32"
"hiking",48,"5.23","5.32","5.15"
"logistics",69,"4.23","*4.42","4.35"
"mystery",10,"21.74","21.74","21.74"
"mystery-prime",10,"22.61","22.9","23.1"
"no-mystery",29,"0","0",""
"parc-printer",1156164,"0","**0",""
"parking",74,"0.18","0.22","0.23"
"pathways",95,"3.99","4.38","4.38"
"peg-solitaire",11,"1.06","**1.06",""
"pipesworld",34,"8.39","8.83","**10.41"
"rovers",49,"6.62","7.33","7.21"
"satellite",37,"7.55","9.08","8.94"
"scanalyzer-3d",62,"2.1","2.1","1.47"
"storage",22,"12.98","13.08","*13.67"
"tetris",119,"1.48","1.81","*5.28"
"thoughtful",75,"3.44","3.49","*3.98"
"tidybot",55,"4.71","4.9","*6.38"
"tpp",81,"4.3","4.46","4.32"
"transport",1682,"2.19","5.58","**6.35"
"trucks",35,"6.07","6.07","6.07"
"woodworking",898,"0.09","0.09","**0.57"
"zenotravel",26,"10.57","*10.96","10.88"
"Total",76,"5.33","5.91","5.85"
\end{filecontents*}
\pgfplotstabletypeset[
    column type=l,
    every head row/.style={
    before row={
    \toprule
    },
    after row=\midrule,
    },
    every last row/.style={before row=\toprule,after row=\midrule},
    display columns/0/.style={column type=l},
    columns/domain/.style ={column name=Domains},
    columns/ceog/.style ={column name=$\mathit{c_{initial}}$, column type=r},
    columns/bj/.style ={column name=$\Delta c_{FIBS\_BJ}$, column type=r},
    columns/fj/.style ={column name=$\Delta c_{FIBS\_GJ}$, column type=r},
    columns/kr/.style ={column name=$\Delta c_{MPR}$, column type=r},
    col sep=comma,
    string type,
     ignore chars={"},
    ]{media/bsd_rr.csv}
    \end{table}
Table \ref{tbl:table2} presents the experimental results of eliminating redundant operators in FIBS using backward justification and greedy justification. We refer to FIBS with backward justification as $FIBS\_BJ$, and with greedy justification as $FIBS\_GJ$. The results are compared with the Minimal Plan Reduction (MPR) approach \cite{Salerno_2023}. For each domain, the column $c_{initial}$ denotes the average cost of the original plan before reduction. The columns $\Delta c_{FIBS\_BJ}$, $\Delta c_{FIBS\_GJ}$, and $\Delta c_{MPR}$ represent the mean (geometric) percentage reduction in plan cost achieved by each method relative to the initial plan cost. We use the geometric mean rather than the arithmetic mean because plan costs vary substantially across domains. The geometric mean provides a fair overall measure, as it reduces the influence of domains with unusually large or small percentage changes.

The effectiveness of the methods varies considerably across domains. Some domains show minimal improvements (e.g., parking 0.18-0.23\%, woodworking 0.09-0.57\%), while others demonstrate substantial cost reductions (e.g., mystery: 21.74\%, mystery-prime: 22-23\%). On average, FIBS\_BJ achieves a geometric mean reduction of 5.33\%, while FIBS\_GJ attains a slightly higher reduction of 5.91\%. On the other hand, the MPR approach yields an overall geometric mean reduction of 5.85\%.

Across all domains, FIBS\_GJ reduces cost equal to or better than FIBS\_BJ. This is because we apply backward justification only on outer blocks in a BDPO plan. This is because an internal block (i.e., one enclosed within another block) may also be backward justified if it produces facts that prevent its parent block from becoming a threat to some causal link \cite{sabah_siddiqui_aic_2024}. FIBS\_GJ outperforms FIBS\_BJ in 21 domains, particularly in barman blocks, child-snack, depot, elevator, pathways, and storage. MPR surpasses FIBS\_GJ significantly in nine domains, notably in barman, depots, pipesworld, storage, transport, and woodworking, whereas FIBS\_GJ outperforms MPR in elevator, genome-edit-distance, logistics, pathways, and zenotravel. The primary reason FIBS\_GJ achieves greater cost reduction in some domains is its ability to minimize plan cost during block substitution, by replacing blocks with lower-cost subplans. Overall, the experimental results demonstrate that integrating justification strategies in BDPO plans provides an effective approach to plan reduction across diverse domains.

\section{Conclusion}
Our method, FIBS, is an iterative, anytime algorithm that continually improves the flexibility of a plan.  Our experimental results show that FIBS significantly improves plan flexibility over EOG and MaxSAT reorderings. The proposed pruning method for redundant operators applied in FIBS also substantially decreases plan cost.  FIBS exploits BDPO plans by considering blocks as candidate subplans for replacement instead of blindly searching for suitable subplans, resulting in lower computational costs. However, other subplans, not enclosed by blocks within a BDPO plan, remain unexplored in our algorithm. One extension of our work involves exploring alternative approaches, such as forming random blocks, to search candidate subplans for substitutions. We can also analyze whether block deordering produces suitable blocks for substitutions by conducting a comparative study with randomized block-substitutions in FIBS.  For further analysis, we can employ planners other than LAMA to generate initial plans, and study the influence of different planners. The concept of block-substitution can be applied to refine plans in other applications, such as assumption-based planning \cite{Davis-Mendelow_2013} and plan quality optimization.
\printbibliography
\end{document}